\documentclass[letterpaper]{article} 
\usepackage[draft]{aaai25}  
\usepackage{times}  
\usepackage{helvet}  
\usepackage{courier}  
\usepackage[hyphens]{url}  
\usepackage{graphicx} 
\usepackage{natbib}  
\usepackage{caption} 
\usepackage{colortbl}
\usepackage{amsmath,amssymb,amsfonts,amsthm}
\usepackage[ruled,linesnumbered]{algorithm2e}
\DeclareMathOperator*{\argmax}{arg\,max}
\DeclareMathOperator*{\argmin}{arg\,min}
\usepackage{booktabs}
\usepackage[dvipsnames]{xcolor}

\newcommand{\expcol}[1]{{#1}}
\newcommand{\unexpcol}[1]{\underline{#1}}
\usepackage{enumitem}
\title{Learning Interpretable Models Using Uncertainty Oracles}
\author{
    Abhishek Ghose,
    Balaraman Ravindran 
}
\affiliations{
    \textsuperscript{\rm 1}Indian Institute of Technology, Madras\\

    abhishek.ghose.82@gmail.com
}

\usepackage{bibentry}

\begin{document}

\maketitle

\begin{abstract}
A desirable property of interpretable models is small size, so that they are easily understandable by humans. This leads to the following challenges: (a) small sizes typically imply diminished accuracy, and (b) bespoke levers provided by model families to restrict size, e.g., L1 regularization, might be insufficient to reach the desired size-accuracy trade-off.

We address these challenges here. Earlier work has shown that learning the training distribution creates accurate small models. Our contribution is a new technique that exploits this idea. The training distribution is encoded as a Dirichlet Process to allow for a flexible number of modes that is learnable from the data. Its parameters are learned using Bayesian Optimization; a design choice that makes the technique applicable to non-differentiable loss functions.  To avoid the challenges with high dimensionality,  the data is first projected down to one-dimension using  uncertainty scores of a separate probabilistic model, that we refer to as the uncertainty oracle.

We show that this technique addresses the above challenges: (a) it arrests the reduction in accuracy that comes from shrinking a model (in some cases we observe $\sim 100\%$ improvement over baselines), and also, (b) that this maybe applied with no change across model families with different notions of size; results are shown for Decision Trees, Linear Probability models and Gradient Boosted Models. 

Additionally, we show that (1) it is more accurate than its predecessor, (2) requires only one hyperparameter to be set in practice, (3) accommodates a multi-variate notion of model size, e.g., both maximum depth of a tree and number of trees in Gradient Boosted Models, and (4) works across different feature spaces between the uncertainty oracle and the interpretable model, e.g., a Gated Recurrent Unit trained using character sequences may be used as an oracle for a Decision Tree that ingests character n-grams as features.

\end{abstract}

\section{Introduction}

In recent years, Machine Learning (ML) models have become increasingly pervasive in various real world systems. This has led to a growing emphasis on models to be \emph{understandable}, especially in domains where the cost of being wrong is prohibitively high, e.g., medicine and healthcare \citep{Caruana:2015:IMH:2783258.2788613,Ustun2016,9233366,jwang2022interpretability}, defence applications \citep{DARPA_XAI,10136827}, law enforcement \citep{COMPAS-bias,COMPAS-recidivism,10.1145/3643834.3661629}.

\begin{figure}[t]
\centering
\includegraphics[width=0.9\columnwidth]{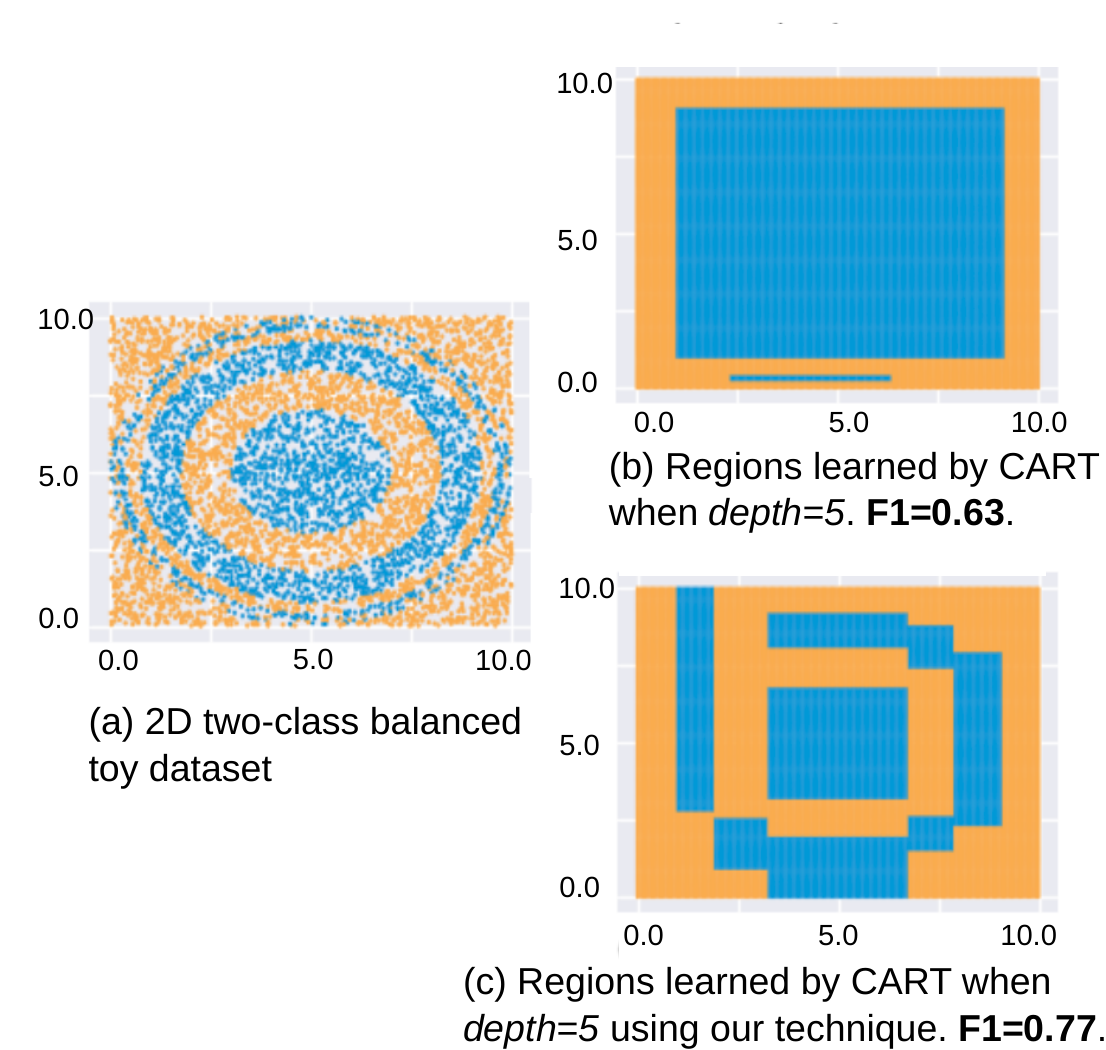}
\caption{Application of our technique is shown on the toy dataset in (a). Learning a DT constrained to a depth of $5$ using the CART algorithm produces the regions shown in (b). Additionally learning the training distribution using our technique produces the regions in (c). For both (b) and (c) the F1-macro scores on a held-out set are reported.}
\label{fig:oracle_demo}
\end{figure}

\begin{figure*}[t]
\centering
\includegraphics[width=1.8\columnwidth]{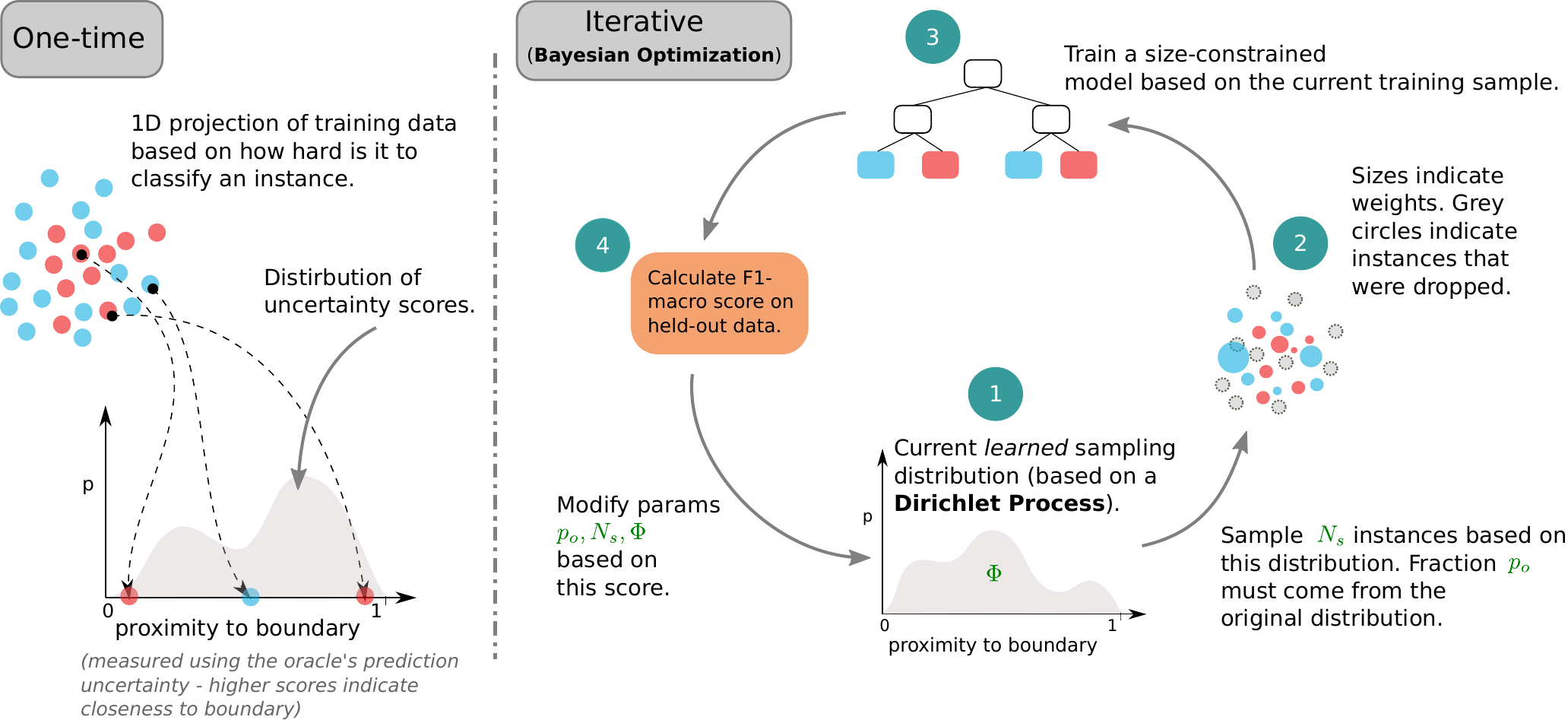}
\caption{Overview of our technique. \emph{Left}: Training instances are characterized by their proximity to class boundaries. As a proxy for this quantity,  we use the prediction uncertainty scores of a probabilistic oracle (these may also be seen as an 1D \emph{projection}): higher uncertainty indicates proximity to a boundary. These scores are calculated \emph{once}. \emph{Right}: The size-constrained model is learned iteratively. A sampling distribution, parameterized by $\Phi$, over the uncertainty values (shown in \texttt{Step 1}) is used to sample training instances (as in \texttt{Step 2}), which is used to train a size-constrained model (shown in \texttt{Step 3}). Its accuracy on a held-out set - \texttt{Step 4} - is used to modify $\Phi$. This loop, \texttt{Steps 1-4}, is executed by a BayesOpt algorithm.}
\label{fig:overall_schematic}
\end{figure*}

An important aspect of model interpretability is its size (smaller is better); this is both shown in various user studies \cite{feldman_boolean_coomplexity,Kulesza2013,PILTAVER2016333,Lage_Chen_He_Narayanan_Kim_Gershman_Doshi-Velez_2019,poursabzi-sangdeh2021manipulating}, and is evidenced by its popularity as an algorithm design criteria \cite{lasso,Ribeiro:2016:WIT:2939672.2939778,DBLP:journals/corr/abs-1711-07414,Lipton:2018:MMI:3236386.3241340,Murdoch22071,Lakkaraju:2016:IDS:2939672.2939874,good2023feature}. However, small size typically implies high bias and thus, relatively lower accuracy. A practitioner may control this size-accuracy trade-off using bespoke levers offered by a training algorithm, e.g., early stopping in Decision Trees (DT), L1 regularization in linear models. This doesn't entirely and conveniently solve the trade-off problem since (1) to get to the desired trade-off, one may need to be intimately aware of how various hyperparameters (hence referred to as \emph{hyperparams}) interact, and (2) the desired trade-off might not even be possible for a model family within its hyperparameter search space.

Here we propose a model-agnostic\footnote{We use the term to mean agnostic to the model \emph{family}, as is accepted usage in the area of XAI, e.g., SHAP \cite{NIPS2017_7062}, LIME \cite{Ribeiro:2016:WIT:2939672.2939778} are considered to be model-agnostic.}  technique that is shown to often produce better accuracies for small-sized models on classification problems. 
The underlying strategy is to learn a distribution over training instances and accordingly create a new training set; this has been shown to produce small accurate models \cite{frontiers_density_tree}. Since a training distribution maybe learned for any setting of a model's hyperparams, the technique maybe seen as a ``meta'' algorithm, where the size constraint is enforced using standard hyperparams, and then accuracy is further improved by learning a distribution over training data. 

We show its application on the toy dataset in Figure \ref{fig:oracle_demo}(a). Figure \ref{fig:oracle_demo}(b) visualizes  class regions learned by a DT of $depth=5$ using the CART \cite{cart93} algorithm. The F1-macro score on a held-out set is $0.63$. When the training distribution is also learned using our technique, we obtain the regions in Figure \ref{fig:oracle_demo}(c) and a F1-macro score of $0.77$.

The workings of the technique itself are presented at a high-level in Figure \ref{fig:overall_schematic}. Instead of learning the training distribution directly, which might be expensive because of the dimensionality of the data, we first project the data down to one dimension. This is done just once, and is shown in the left panel in Figure \ref{fig:overall_schematic}.  Since we are solving for classification, we pick this dimension to be a numeric indicator of how close an instance is to a class boundary. As a convenient proxy, we train a separate highly accurate probabilistic \emph{oracle} model on the training data\footnote{In Figure \ref{fig:oracle_demo}, the oracle used is a Gradient Boosted Model.}, and use its prediction \emph{uncertainty} score as the projected value; higher uncertainty scores ideally denote greater proximity to class boundaries.

The distribution is modeled as an \emph{Infinite Beta Mixture Model} using a \emph{Dirichlet Process}, which is \emph{iteratively} learned. \texttt{Step 1} on the right panel in Figure \ref{fig:overall_schematic} shows the current distribution, based on which training data is sampled (\texttt{Step 2}). The size-constrained model of interest is then trained on this sample - \texttt{Step 3} - and its accuracy on a held-out set is calculated - \texttt{Step 4}. This score is used as a feedback for the optimizer which repeats the process to learn better distribution parameters. We use \emph{Bayesian Optimization (BayesOpt)} \cite{7352306,garnett_bayesoptbook_2023} to accommodate models with non-differentiable losses. 

\qed

\emph{Our contributions} in this work are: we propose a technique that (a) produces small accurate models by learning a training distribution. It is shown to provide relative improvements of $\sim 100\%$ in some cases. (b) is model-agnostic, where the models may also have non-differentiable losses, e.g., DTs. Additionally, we show that: (1) it is more accurate than its predecessor \cite{frontiers_density_tree},  (2) is practically convenient since only one hyperparameter needs to be set, (3) allows for model sizes to be defined by more than one component, e.g., number of trees and depth per tree in Gradient Boosted Models (GBM) \cite{gbm}, and (4) can use oracles trained on a different feature representation of the same data. The last property is useful since it allows for a broad choice of oracles. We show an example of this later. 

The rest of the paper is organized as follows: we first review related work in \S \ref{sec:related_work}. We then detail our technique in \S \ref{sec:methodology}. We follow that up with rigorous empirical validations in \S \ref{sec:val} and comparisons to prior work in \S \ref{sec:compare_results}. After discussing additional applications, in \S \ref{sec:vector_model_sizes} and \S \ref{sec:different_feature_space}, we conclude the paper with a summary and notes on future work in \S \ref{sec:conclusion}.

\section{Related Work}
\label{sec:related_work}
The concept of using a different training distribution relative to test is common in the case of class imbalance, e.g., undersample the majority class data \cite{Japkowicz:2002:CIP:1293951.1293954,SMOTE,ADASYN,10.1145/3152494.3152496}, but it was shown to be a general strategy for improving accuracy in \citet{frontiers_density_tree}. Their technique relies on a specialized DT called \emph{density trees}, that encodes the geometric placement of training data, and facilitates learning a distribution. In our view, notwithstanding its interesting properties,  the accuracy of density trees implicitly limit the effectiveness of their technique. This work may be seen as a non-trivial extension: a model (the uncertainty oracle) from an arbitrary model family may be used, resulting in greater flexibility and accuracy.

The interaction of two models - the oracle and the interpretable model - might suggest an overlap with the area of \emph{Knowledge Distillation} \cite{kd_survey_Gou2021}. But there is a critical difference: \emph{in theory, we don't require the oracle model}; its sole purpose here is to reduce computational cost. Its a dimensionality reduction tool\footnote{There are others possible, e.g., given point $A$, use the number of points of different classes among its nearest neighbors to estimate $A$'s proximity to a class boundary.} that is invoked once outside the main learning loop (see Figure \ref{fig:overall_schematic}). This peripheral role is emphasized by the fact that the labels predicted by the oracle are ignored.

For similar reasons, this technique must not be seen as a successor to model explanation techniques where we seek to explain the oracle, e.g.,  \citet{trepan,NEURIPS2019_567b8f5f}; again, the key difference is that any form of alignment/fidelity wrt the oracle is ignored. 

Since \citet{frontiers_density_tree} represents the work closest to ours, we benchmark our technique against their approach.

\section{Methodology}
\label{sec:methodology}
We describe our technique in detail in this section. But first, we introduce some notation.
\subsection{Terminology and Notation}
\begin{enumerate}
    \item A dataset  is denoted as a set of instance-label pairs, $D=\{(x_1, y_1), (x_2, y_2), ..., (x_N, y_N)\}$. A joint distribution over a dataset is denoted by $p(X,Y)$.
    \item To distinguish between the distribution we are given (in form of the dataset) and the one we learn, we refer to the former as the \emph{original} distribution. In \emph{all} experiments here, the test and held-out data follow the original distribution; for the training data, we learn a new distribution.
    \item We let $acc(M, p)$ denote the classification accuracy of model $M$ on data represented by the joint distribution $p(X, Y)$. Here $acc$ is a generic metric, and may be implemented as \emph{AUC}, \emph{F1-macro} score, etc.
    
    \item $train_{\mathcal{F},f}(p, \eta)$ is understood to produce a model of size $\eta$ (for some pre-decided notion of size) from the model family $\mathcal{F}$ using a specific training algorithm $f$. 
    
    For instance, $\mathcal{F}$ might represent DTs and $f$ might be the CART algorithm, and $\eta=5$ might denote a DT of $depth=5$. We let $\eta = *$ denote unbounded size.
\end{enumerate}
Let us state our objective using this notation. Typically, a model is trained on the same distribution as the test (on which it is evaluated), i.e., we evaluate $acc(train_{\mathcal{F},f}(p, \eta), p)$. Here, the training distribution is allowed to be different relative to the test. In other words, we seek $p'$ s.t.:
\begin{equation}
\argmax_{p'} acc(train_{\mathcal{F},f}(p', \eta), p)    
\end{equation}

\subsection{Algorithm}
Referring to the high-level flow in Figure \ref{fig:overall_schematic}, we note that the proposed technique relies on a few important ingredients. These are discussed below:
\begin{enumerate}
    \item \textbf{Uncertainty score}: This is needed for the one-time projection using the oracle. There are multiple ways to measure prediction uncertainty; here we choose \emph{margin uncertainty} \cite{Scheffer:2001:AHM:647967.741626}, since (a) it accounts for prediction probabilities of different classes, (b) while also producing high scores even with two dominant predicted classes in a setting that has more classes. The uncertainty score for $x$, as provided by model $M$, is denoted by $u_M(x) \in [0, 1]$. The margin uncertainty is calculated as:

    \begin{equation}
    u_M(x) \gets 1 - (p_{C_1} - p_{C_2})
    \end{equation}

    Here, $p_{C_1}$ and $p_{C_2}$ denote the probabilities of the most confident and next most confident classes. See \S \ref{sec:appendix_unc_metrics} for further details. 
    
    \item \textbf{Density model}: Since we want to \emph{learn} a distribution, we want the representation to be flexible. We encode the density as a \emph{mixture model} of \emph{Beta} distributions. We use the latter since their support matches the range of uncertainty scores, i.e.,  $u_M(x) \in [0, 1]$. We also note that a \emph{Beta} mixture model can approximate any distribution in $[0, 1]$ arbitrarily well given a sufficient number of components \citep{beta_diaconis}. Further, in the interest of flexibility, we refrain from explicitly dictating the number of \emph{Beta} components, and thus, we use an \emph{Infinite Beta Mixture Model (IBMM)}, where the component assignments are decided by a standard \emph{Dirichlet Process (DP)} \cite{dirichletprocess}. Another advantage of this formulation is that it leads to a fixed number of parameters irrespective of the number of active components\footnote{Contrast this with a \emph{Gaussian Mixture Model (GMM)} where the number of parameters change with the number of components. GMMs have a \emph{conditional} parameter space, and most optimizers handle fixed parameter spaces.}, which makes it easy to pick an optimizer. We note here that \citet{frontiers_density_tree} also use a DP-based IBMM, but for modeling the height of density trees. 

    Two sets of parameters are required to describe this density model: 
    \begin{enumerate}
        \item The shape parameters $A_i, B_i$ of the $i^{th }$ \emph{Beta} component.  These are separately sampled from prior distributions that are themselves \emph{Beta} distributions, with shape parameters $a, b$ and $a', b'$ respectively. Since naively doing this would restrict $A_i$ or $B_i$ to \emph{Beta's} support, i.e., $[0,1]$, we also multiply the sampled value by a variable $scale$, that we set to be large enough to cover the family of component distributions we require\footnote{\textbf{NOTE}: This is fixed at a value of $10000$ and not learned; hence it isn't counted as a parameter.}. Effectively then, $A_{i} \sim scale \times Beta(a, b)$ and $B_{i} \sim scale \times Beta(a', b')$.
        \item The DP needs just a \emph{concentration} parameter $\alpha \in \mathbb{R}_{>0}$ that decides the number of active components, i.e., ones with instances assigned to them\footnote{Of course, in theory, there are an infinite number of components, and the number of active components grows with data.}.
    \end{enumerate}
    In all, the density model requires \emph{five} parameters,  which we denote as $\Psi=\{\alpha, a, b, a', b'\}$. To sample $N_s$ instances given $\Psi$, we first determine the number of instances per component using a standard technique like the \emph{Chinese Restaurant Process} \cite{ref10.1007/BFb0099421} and then sample component-wise. Please see \S \ref{sec:sampling_ibmm} for details.

    \item \textbf{Optimization}: As mentioned earlier, we use BayesOpt to accommodate non-differentiable losses. Its resilience to noise is also beneficial since observed model accuracies may be noisy, due to randomized initialization of model parameters, different dataset  splits across trials, etc. Specifically, we use the \emph{hyperopt} library \cite{Bergstra:2013:MSM:3042817.3042832}, which implements the \emph{Tree Structured Parzen Estimator (TPE)} algorithm \cite{Bergstra:2011:AHO:2986459.2986743} . 
    
    For optimization, in addition to $\Psi$, we retain the following parameters originally introduced in \citet{frontiers_density_tree}:
    \begin{enumerate}
        \item $N_s$: This is the sample size - this is also learned.
        \item $p_o \in [0,1]$: Proportion of the new training sample that is constituted by a uniform sample from the original training data. This serves two purposes: (1) it acts as a ``shortcut'' for the optimizer to mix in the original distribution as needed, and  (2) it serves as a ``probe variable'', i.e., it shows how much of the original distribution is actually needed for good accuracies.
    \end{enumerate}
    Accounting for these, we now have a total of \emph{seven} optimization variables: $\Psi=\{\alpha, a, b, a', b'\}, p_o, N_s$, which are iteratively optimized, till the budgeted number of iterations, $T$, are exhausted. These variables are collectively denoted as $\Phi=\{\Psi, N_s, p_o\}$. Algorithm \ref{algo:main_with_data} outlines the overall technique; here the interpretable and oracle model families are denoted by $\mathcal{I}$ and  $\mathcal{O}$, and the respective training algorithms are denoted by $h$ and $g$ respectively. \S \ref{sec:notes_algorithm} provides additional details around model selection, robust estimation of $acc$, etc.
\end{enumerate}

\textbf{Optimization variables and parameters}: The task of the optimizer is to find $\Phi$ that maximizes the held-out accuracy  (line 11 in Algorithm \ref{algo:main_with_data}) within $T$ iterations. The optimizer here accepts \emph{box constraints}, and as such their lower/upper bounds, which need to be set by the user, are \emph{parameters} (along with $T$) of the technique. We discuss in \S \ref{sec:default_params} that reasonable defaults exist for  parameters $\Phi$, e.g., its easy to see $p_o \in [0,1]$. So, in practice $T$ is the only parameter that a user needs to determine. 

\textbf{Smoothing}: A final practical consideration is the smoothness of the optimization landscape. Uncertainty scores over the training data may often result in a density that isn't smooth, making it difficult for the optimizer to identify a good maxima. We redress this by explicitly smoothing the density. We detail this in \S \ref{ssec:smoothe_opt}.

\begin{algorithm}[tb]
 \KwData{Dataset $D$, model size $\eta$, $train_{\mathcal{O}, h}()$, $train_{\mathcal{I}, g}()$, iterations $T$}
 \KwResult{Optimal parameters $\Phi^*$, test set accuracy $s_{test}$ at $\Phi^*$, and interpretable model $M^*$ at $\Phi^*$}
Create splits $D_{train}, D_{val}, D_{test}$ from $D$, stratified wrt labels. Here $|D_{train}|: |D_{val}|: |D_{test}|:: 60:20:20$.\\
$M_O \gets train_{\mathcal{O},h}(D_{train}, *)$ \\

\For {$t\gets1$ \KwTo $T$ } { 
   $\Phi_{t} \gets suggest(s_0, s_1, ... s_{t-1}, \Phi_0, \Phi_1, ..., \Phi_{t-1})$ \tcp{$s_0, \Phi_{0}$ initialized at $t=0$, see text. Note: $\Phi_t = \{\Psi_t, N_{s,t}, p_{o,t}\}$ where $\Psi_t = \{\alpha_t, a_t, b_t, a'_t, b'_t\}$.}

$N_o \gets p_{o,t} \times N_{s,t}$  \\
$N_u \gets N_{s\_t} - N_o$ \\
$D_o \gets$ uniformly sample with replacement  $N_o$ points from  $D_{train}$\\ 
$D_u \gets$ sample $N_u$ points from $D_{train}$  using the DP-based IBMM given current values for $N_u, M_O, D_{train}, \Psi_t$ \tcp{see Algorithm \ref{sec:sampling_ibmm} for details}
  $D_s \gets D_o \uplus D_u$ \tcp{$D_o$, $D_u$ are assumed to be multisets}
  $M_t \gets train_{\mathcal{I}, g}(D_s, \eta) $\\
  $s_t \gets acc(M_t, D_{val})$
 }
$t^* \gets \argmax_t{\{s_1, s_2, ..., s_{T-1}, s_T\}}$\\
$\Phi^* \gets \Phi_{t^*}$\\
$M^* \gets M_{t^*}$\\
$s_{test} \gets acc(M^*, D_{test})$\\
\Return $\Phi^*$, $s_{test}$, $M^*$
 \caption{Learning interpretable model using oracle}
 \label{algo:main_with_data}
\end{algorithm}


This  concludes our discussion of algorithmic details; next, we look at empirical validation.

\section{Experiments}
\label{sec:experiments}
This section covers the various empirical investigations. Some common elements across our experiments are:
\begin{enumerate}
    \item Datasets: We use the following $13$ publicly available datasets for our experiments: \emph{cod-rna, ijcnn1, higgs, covtype.binary, phishing, a1a, pendigits, letter, Sensorless, senseit\_aco, senseit\_sei, covtype, connect-4}. These were obtained from the LIBSVM website \citep{CC01a}. For details, such as number of classes and extent of imbalance, please see \S \ref{sec:datsets}.

    $10000$ instances from each dataset are used. The split ratio used in Algorithm \ref{algo:main_with_data} is $|D_{train}|: |D_{val}|: |D_{test}|:: 60:20:20$, where the splits are stratified wrt labels.
    
    \item Interpretable model families: we use \emph{Linear Probability Models (LPM)}\footnote{We have not used the more common \emph{Logistic Regression} because: (1) LPMs are considered more interpretable \citep{Mood291880}, and (2) LPMs results are indicative of behavior of linear models in general.} and the DTs (produced by the CART algorithm). The notion of model size for LPMs is the number of non-zero coefficients, and  sizes $\eta \in \{1,2, ..., 15\}$  are explored (except for \emph{cod-rna}, that has $8$ features, and so we cannot have a sizes greater than $8$). 
    
    For DTs, the notion of size is depth. For a dataset, we first learn a tree (with no size constraints) with the highest \emph{F1-macro} score using standard $5-$fold cross-validation. We refer to this as the optimal tree $T_{opt}$, and its depth as  $depth(T_{opt})$. We then explore model sizes $\eta \in \{1,2, ..., min(depth(T_{opt}), 15)\}$. 
    Stopping early makes sense since the model is saturated in its learning at $depth(T_{opt})$; changing the input distribution is not helpful beyond this point.
    
    \item Oracle families: As oracles we use \emph{Random Forests (RF)} \cite{randomforests} and \emph{GBMs} \cite{gbm}. Both oracles were calibrated \citep{Platt99probabilisticoutputs,Niculescu-Mizil:2005:OCP:3020336.3020388} for reliable probability estimates.
    
    \item Optimization budget: For DTs, we use $T=3000$, while for LPMs $T=1000$ is used. These values were determined based on limited search. The budget for LPMs is lower since for multi-class datasets ($7$ of $13$ here) we construct one-vs-all models which makes training LPMs time-consuming.
\end{enumerate}

Our implementation primarily utilizes the following libraries: \emph{scikit-learn} \cite{scikit-learn}, \emph{scipy} \cite{scipy}, \emph{LightGBM} \cite{Ke:2017:LHE:3294996.3295074}.

First, we look at the validation results.

\subsection{Validation}
\label{sec:val}
In these experiments, our goal is to scrutinize how our technique performs. 
For various combinations of models and oracles, i.e., $\{LPM, DT\} \times \{GBM, RF\}$, we measure the percentage relative improvement in the \emph{F1-macro} score (on $D_{test}$) in terms of the baseline score $F1_{test}^{base}$ and the one produced by our model, $F1_{test}^*$:
\begin{equation}
    \label{eqn:delta_f1}
    \delta F1_{test} = \frac{100 \times (F1_{test}^* - F1_{test}^{base})}{F1_{test}^{base}}
\end{equation}
We use the macro score since its not impacted by class imbalance. 

In the interest of robustness we run \emph{five trials} per configuration, i.e., a combination of dataset, oracle family, model family and size, and utilize the validation set to accept the model produced by our technique $M^*$. Specifically: indexing trials with $i$, we conduct an independent \emph{t-test} on $\{F1_{val}^*\}_{1\leq i \leq 5}$ and $\{F1_{val}^{base}\}_{1\leq i \leq 5}$. The null hypothesis is that $M^*$ doesn't produce results different $M^{base}$. If we can reject the null at a significance of $p=0.1$, we report $\delta F1_{test}$ as in Equation \ref{eqn:delta_f1}\footnote{The test scores from different trials are averaged first.}, else we report $\delta F1_{test}=0$, i.e., we reject $M^*$. Here $\delta F1_{test} \in (-\infty, \infty)$; negative values are possible since we  pick a model based on $D_{val}$ while we report based on $D_{test}$.

Table \ref{tab:internal_improvements_short} shows a portion of the results in the interest of space - for complete results (and analysis of statistical significance), please see \S \ref{sec:appendix_val_results}.

\begin{table*}[hbt!] \small
            \centering
            \caption{\small This table shows the average improvements, $\delta F1$, over \textbf{five runs} for the combinations \textbf{model=\emph{\{LPM, DT}\}} and \textbf{oracle=\emph{GBM}}.  The improvements are measured relative to the model at the first iteration. Here, $\delta F1 \in (-\infty, \infty)$. Negative improvements are shown in \unexpcol{underlined}. \textbf{Complete results, including analysis of statistical significance, are presented in \S \ref{sec:appendix_val_results}}.}
            \label{tab:internal_improvements_short}
            \hspace{0.5pt}
            \resizebox{\textwidth}{!}{\begin{tabular}{lllllllllllllllll}
            \toprule [1.1pt]
dataset & model\_ora & 1 & 2 & 3 & 4 & 5 & 6 & 7 & 8 & 9 & 10 & 11 & 12 & 13 & 14 & 15\\
 \toprule [1.1pt]
cod-rna & lpm\_gbm & 1.39 & 12.53 & {14.76} & {15.73} & 14.97 & 12.00 & 0.00 & {0.08} & - & - & - & - & - & - & - \\  
 & dt\_gbm & {0.00} & {0.00} & 0.00 & 1.26 & 0.00 & {0.00} & 0.00 & 0.00 & \unexpcol{-0.28} & 0.08 & - & - & - & - & - \\  
\midrule
ijcnn1 & lpm\_gbm & \unexpcol{-0.16} & {3.36} & {3.93} & 0.00 & {5.19} & {4.18} & {3.85} & {3.79} & {3.69} & 2.99 & {2.97} & 3.21 & 3.11 & 3.26 & 3.02 \\  
 & dt\_gbm & 1.96 & 12.00 & {10.15} & {11.37} & {10.63} & 7.18 & 3.63 & {4.52} & {2.91} & {1.78} & {1.93} & {2.29} & 1.47 & {2.26} & 0.00 \\  
\midrule
higgs & lpm\_gbm & {29.29} & {17.80} & 11.40 & 6.56 & 3.06 & 2.68 & {3.16} & {2.90} & {2.67} & {2.82} & 2.65 & 1.79 & 2.62 & 2.19 & 1.63 \\  
 & dt\_gbm & 0.00 & 0.00 & {1.86} & 0.26 & 0.93 & 0.45 & - & - & - & - & - & - & - & - & - \\  
\midrule
covtype.binary & lpm\_gbm & 76.52 & {66.39} & {29.17} & {12.51} & {9.18} & {5.28} & {4.94} & {4.56} & {3.92} & {3.56} & {3.62} & {3.31} & {2.59} & {2.83} & {2.39} \\  
 & dt\_gbm & {0.00} & {0.00} & {2.35} & 1.27 & 1.18 & 1.11 & 0.00 & 0.00 & 0.00 & - & - & - & - & - & - \\  
\midrule
phishing & lpm\_gbm & {0.00} & 1.88 & 2.88 & 3.05 & 3.22 & 3.25 & 2.99 & 1.69 & 1.42 & {1.45} & {1.29} & 0.00 & 0.00 & 0.00 & 0.00 \\  
 & dt\_gbm & {0.00} & 0.00 & {0.00} & 0.07 & {0.39} & {0.00} & {0.28} & 0.22 & {0.44} & {0.23} & 0.00 & {0.00} & {0.00} & 0.00 & {0.00} \\  
\midrule
a1a & lpm\_gbm & {0.00} & 2.55 & 7.58 & 8.98 & 8.40 & 8.03 & 8.90 & 8.23 & 8.17 & 7.90 & 5.96 & 7.10 & 6.97 & 6.18 & 5.73 \\  
 & dt\_gbm & {0.00} & 5.54 & 2.39 & 3.84 & {3.55} & 2.55 & 1.51 & 2.25 & 4.87 & - & - & - & - & - & - \\  
\midrule
pendigits & lpm\_gbm & {51.39} & {23.44} & 16.18 & {8.95} & {8.84} & {6.63} & 4.86 & {1.83} & 2.27 & 2.16 & {2.44} & 2.16 & {3.33} & 2.97 & {2.73} \\  
 & dt\_gbm & 14.02 & {6.72} & 5.11 & 13.14 & 6.42 & 4.20 & 2.46 & {1.09} & {0.98} & {0.16} & \unexpcol{-0.26} & {0.00} & {0.00} & {0.00} & {0.00} \\  
\midrule
letter & lpm\_gbm & 57.06 & 48.48 & 59.85 & {29.76} & {36.09} & 19.27 & 20.37 & 16.08 & 17.55 & 15.16 & 17.26 & 16.51 & 18.46 & 17.19 & 15.55 \\  
 & dt\_gbm & {0.00} & {13.98} & 25.05 & {33.96} & 32.05 & 15.49 & {11.17} & {0.00} & {4.26} & {3.50} & {1.99} & 0.00 & 0.00 & 0.00 & {0.00} \\  
\midrule
Sensorless & lpm\_gbm & 216.47 & {257.56} & {178.31} & {117.01} & {90.70} & {83.90} & {73.50} & {65.95} & 61.57 & 57.97 & 56.54 & 57.15 & 55.45 & 66.24 & 68.24 \\  
 & dt\_gbm & \unexpcol{-0.01} & 42.42 & {68.13} & 44.38 & {17.39} & {10.32} & 1.82 & {1.44} & {0.79} & {0.64} & 0.41 & 0.12 & {0.00} & \unexpcol{-0.02} & {0.34} \\  
\midrule
senseit\_aco & lpm\_gbm & 173.71 & 170.68 & 63.95 & {44.20} & 33.49 & 22.99 & 19.14 & 13.50 & 10.29 & 7.59 & 6.26 & 5.92 & {5.30} & 4.89 & 4.32 \\  
 & dt\_gbm & 14.89 & 0.00 & {3.71} & 2.32 & {4.85} & 0.81 & {0.00} & - & - & - & - & - & - & - & - \\  
\midrule
senseit\_sei & lpm\_gbm & 160.59 & {65.27} & 23.44 & 10.48 & 6.76 & 4.86 & 4.82 & 4.46 & 4.79 & 4.12 & 4.54 & {5.17} & 3.91 & 4.21 & {4.46} \\  
 & dt\_gbm & {2.66} & {1.01} & {3.49} & {2.29} & {0.95} & {1.30} & {1.37} & 0.00 & - & - & - & - & - & - & - \\  
\midrule
covtype & lpm\_gbm & {36.87} & {49.24} & {12.78} & {11.21} & 7.84 & 7.15 & 7.15 & 8.07 & 7.70 & 8.25 & {10.94} & {8.35} & 4.37 & 8.77 & 5.84 \\  
 & dt\_gbm & 342.27 & 92.85 & 43.23 & {20.04} & 8.14 & {8.05} & {5.67} & 3.26 & {4.92} & {3.52} & {2.72} & 0.00 & {0.00} & {0.00} & {1.74} \\  
\midrule [0.5pt]
connect-4 & lpm\_gbm & {37.62} & 11.66 & 12.01 & 6.84 & 5.68 & 6.82 & 4.58 & 2.10 & 3.82 & 3.21 & {3.02} & {3.64} & {2.32} & {2.97} & {3.40} \\  
 & dt\_gbm & 89.33 & {29.23} & 20.20 & {12.10} & 9.73 & 9.88 & 7.82 & 7.43 & 0.57 & 4.61 & 1.08 & {3.35} & 2.23 & {1.15} & {1.55} \\

\bottomrule [1.1pt]
\end{tabular}} \end{table*}

\emph{Observations}: We highlight some interesting trends:
\begin{enumerate}
    \item We note that the incidence of negative improvements if fairly low. Of course, this result set is incomplete, but referring to the complete set in \S \ref{sec:appendix_val_results}, we note that only $13$ of $690$ non-null observations, or $1.88\%$, are negative. The average negative improvement is $-0.24\%$.
    
    \item As model size increases (left to right in Table \ref{tab:internal_improvements_short}), positive improvements (which can be high for small sizes, e.g., $> 100\%$) tend to reduce. This makes intuitive sense since beyond a certain model size, when all informative patterns in the data have been captured, modifying the training distribution should not have much/any effect.
    
    \item For DTs, the drop in improvements happen earlier than for LPMs. An intuitive explanation for this is that an unit increase in size for the LPM and DT do not lead to identical increase in capacity. DTs are non-linear models to begin with, and then, increasing their depth by one leads to a much larger increment in capacity, e.g., it doubles the number of leaves for a binary tree.
\end{enumerate}

Points \#2 and \#3 above are similar to trends reported in \citet{frontiers_density_tree}, and thus they informally validate that these techniques are at least qualitatively similar.

\subsection{Comparison to Previous Work}
\label{sec:compare_results}
In this section, we compare our technique to the one based on density trees. Please see \S \ref{sec:related_work} for why we consider this to be the incumbent.

Their metric is slightly different from ours. Instead of reporting results for $F1^*$, they report them for $max(F1^*, F1^{base})$. This is an ``outcome-centric'' view\footnote{Another reason provided is that with a sufficient budget the optimizer will eventually learn to set $p_o=1$, thus emulating  $M^{base}$ exactly, if $M^{base}$ is indeed the best possible model. In this case $\delta F1=0$ as per Equation \ref{eqn:delta_f1}.}, where you can't do worse than your best model. For this case, $\delta F1_{test} \in [0, \infty) $. We also follow this scoring scheme in this section to match their reporting.

We report two scores for comparison:
\begin{enumerate}
    \item To compare improvements, we use the \emph{Scaled Difference in Improvement (SDI)}:

    \begin{align}
    \label{eqn:define_sdi}
    SDI &= 
    \begin{cases}
    (\delta F1^{ora} - \delta F1^{den})/H, &\text{ if } H>0\\
     0, &\text{ if } H=0  \\
     \end{cases}\\
   & \text{where } H=\max{\{\delta F1^{den}, \delta F1^{ora}\}} \nonumber
\end{align}

    Here $\delta F1^{ora}$  and $\delta F1^{den}$ are the improvements from our technique and by using density trees, respectively. The scaling wrt $H$ ensures that $SDI\in [-1, 1]$ making it convenient to interpret. Note that $H \geq 0$ since both $\delta F1^{ora} \geq 0$ and $\delta F1^{den} \geq 0$ in the current scoring scheme. For brevity, we average the $SDI$ scores at the level of a dataset, across model sizes, for a given model and oracle. This averaged score is denoted by $\overline{SDI}$, and this is what we report.

    \item Since  $\overline{SDI}$ is aggregated over model sizes, we also report the percentage of times $\delta F1^{ora} > \delta F1^{den}$ across these model sizes. This is denoted as $pct\_better$

\end{enumerate}

All  $\delta F1^{ora}$ and $\delta F1^{den}$ scores used are the \emph{averaged over five runs}. 

We consider our approach to be better if $\overline{SDI} \boldsymbol{> 0}$ \emph{and} $pct\_better \boldsymbol{> 50\%}$. These scores are shown in Table \ref{tab:dentr_LPM}. Since the density trees approach lacks a notion of an oracle, we present results for GBMs and RFs separately. Numbers that represent superior performance by density trees are underlined. Note also the two special groupings:
    \begin{itemize}
        \item \textbf{ANY}: For each model size, the $SDI$ score considered is the higher of the ones obtained from using the $GBM$ or $RF$ as oracles. The $\overline{SDI}$ and $pct\_better$ scores are computed based on these scores. This grouping represents the ideal way to use our technique in practice: try multiple oracles and pick the best.
        \item \textbf{OVERALL}: This averages results across datasets, to provide an aggregated view.
    \end{itemize}
    The cells identified by \textbf{OVERALL} \emph{and}  \textbf{ANY} provide comparison numbers aggregated over datasets, model sizes and oracles.


\begin{table*}[!htbp] \small
                    \centering
                    \caption{\small LPM, DT compared to the Density Tree approach. All  $\delta F1^{ora}$ and $\delta F1^{den}$ scores used are the \emph{average over five runs}. Cases where density trees fare better are \underline{underlined}. The line in the middle separates binary class datasets (top) from multi-class ones (bottom).}\label{tab:dentr_LPM}
                    \hspace{0.5pt}
\resizebox{0.85\textwidth}{!}{\begin{tabular}{lrrrrrr}
\toprule
\multicolumn{1}{r}{}& \multicolumn{3}{c}{LPM} & \multicolumn{3}{c}{DT} \\ 
\cmidrule(lr){2-4}
\cmidrule{5-7}
dataset & \multicolumn{1}{c}{GBM} & \multicolumn{1}{c}{RF}  & \multicolumn{1}{c}{\textbf{ANY}} & \multicolumn{1}{c}{GBM} & \multicolumn{1}{c}{RF}  & \multicolumn{1}{c}{\textbf{ANY}} \\
\toprule
cod-rna & 	\unexpcol{-0.38}, \unexpcol{  0.00\%} & \unexpcol{-0.45}, \unexpcol{  0.00\%} & \unexpcol{-0.33}, \unexpcol{  0.00\%} & \expcol{0.51}, \expcol{ 60.00\%} & \expcol{0.50}, \expcol{ 70.00\%} & \expcol{0.65}, \expcol{ 80.00\%}  \\
ijcnn1 & 	\expcol{0.06}, \expcol{ 66.67\%} & \expcol{0.11}, \expcol{ 80.00\%} & \expcol{0.20}, \expcol{ 93.33\%} & \expcol{0.23}, \expcol{ 53.33\%} & \expcol{0.68}, \expcol{100.00\%} & \expcol{0.68}, \expcol{100.00\%}  \\
higgs & 	\unexpcol{-0.07}, \unexpcol{ 40.00\%} & \unexpcol{-0.07}, \unexpcol{ 40.00\%} & \expcol{0.04}, \unexpcol{ 46.67\%} & \expcol{0.23},  50.00\% & \expcol{0.61}, \expcol{ 83.33\%} & \expcol{0.61}, \expcol{ 83.33\%}  \\
covtype.binary & 	\unexpcol{-0.16}, \unexpcol{ 40.00\%} & \unexpcol{-0.33}, \unexpcol{ 13.33\%} & \unexpcol{-0.15}, \unexpcol{ 40.00\%} & \expcol{0.23}, \expcol{ 66.67\%} & \expcol{0.26}, \expcol{ 72.73\%} & \expcol{0.38}, \expcol{ 81.82\%}  \\
phishing & 	\expcol{0.30}, \expcol{ 80.00\%} & \expcol{0.37}, \expcol{ 86.67\%} & \expcol{0.38}, \expcol{ 86.67\%} & \expcol{0.11}, \unexpcol{ 26.67\%} & \unexpcol{-0.00}, \unexpcol{ 26.67\%} & \expcol{0.23}, \unexpcol{ 46.67\%}  \\
a1a & 	\unexpcol{-0.03}, \expcol{ 60.00\%} & \expcol{0.13}, \expcol{ 66.67\%} & \expcol{0.13}, \expcol{ 66.67\%} & \unexpcol{-0.06}, \unexpcol{ 44.44\%} & \expcol{0.43}, \expcol{ 75.00\%} & \expcol{0.52}, \expcol{ 83.33\%}  \\ \midrule
pendigits & 	\expcol{0.59}, \expcol{100.00\%} & \expcol{0.59}, \expcol{ 93.33\%} & \expcol{0.62}, \expcol{100.00\%} & \expcol{0.23}, \expcol{ 60.00\%} & \expcol{0.16}, \unexpcol{ 46.67\%} & \expcol{0.25}, \expcol{ 60.00\%}  \\
letter & 	\expcol{0.79}, \expcol{100.00\%} & \expcol{0.81}, \expcol{100.00\%} & \expcol{0.81}, \expcol{100.00\%} & \expcol{0.02}, \unexpcol{ 33.33\%} & \unexpcol{-0.34}, \unexpcol{ 13.33\%} & \expcol{0.06}, \unexpcol{ 40.00\%}  \\
Sensorless & 	\expcol{0.64}, \expcol{100.00\%} & \expcol{0.65}, \expcol{100.00\%} & \expcol{0.66}, \expcol{100.00\%} & \unexpcol{-0.23}, \unexpcol{ 20.00\%} & \unexpcol{-0.39}, \unexpcol{ 20.00\%} & \unexpcol{-0.23}, \unexpcol{ 20.00\%}  \\
senseit\_aco & 	\expcol{0.55}, \expcol{100.00\%} & \expcol{0.63}, \expcol{100.00\%} & \expcol{0.63}, \expcol{100.00\%} & \expcol{0.50}, \expcol{ 85.71\%} & \expcol{0.37}, \expcol{ 75.00\%} & \expcol{0.39}, \expcol{ 75.00\%}  \\
senseit\_sei & 	\expcol{0.61}, \expcol{100.00\%} & \expcol{0.66}, \expcol{100.00\%} & \expcol{0.67}, \expcol{100.00\%} & \unexpcol{-0.25}, \unexpcol{ 42.86\%} & \expcol{0.51}, \expcol{100.00\%} & \expcol{0.51}, \expcol{100.00\%}  \\
covtype & 	\expcol{0.20}, \expcol{ 80.00\%} & \expcol{0.39}, \expcol{ 93.33\%} & \expcol{0.43}, \expcol{100.00\%} & \expcol{0.26}, \expcol{ 66.67\%} & \expcol{0.16}, \expcol{ 66.67\%} & \expcol{0.40}, \expcol{ 80.00\%}  \\
connect-4 & 	\expcol{0.23}, \expcol{ 73.33\%} & \expcol{0.24}, \expcol{ 66.67\%} & \expcol{0.38}, \expcol{ 86.67\%} & \unexpcol{-0.23}, \unexpcol{ 33.33\%} & \unexpcol{-0.13}, \expcol{ 53.33\%} & \expcol{0.08}, \expcol{ 66.67\%}  \\
\textbf{OVERALL} & 	\expcol{0.28}, \expcol{ 75.00\%} & \expcol{0.32}, \expcol{ 75.00\%} & \expcol{0.37}, \expcol{ 81.38\%} & \expcol{0.10}, \unexpcol{ 47.06\%} & \expcol{0.16}, \expcol{ 57.23\%} & \expcol{0.31}, \expcol{ 67.30\%}  \\

\bottomrule 
\end{tabular}} \end{table*}

The predominance of non-underlined values indicate that our technique performs better in most settings. In both cases, the \textbf{OVERALL} $+$\textbf{ANY} entries indicate that our technique works better on average - in terms of both the extent of improvement $\overline{SDI}$ and its frequency $pct\_better$. 

\section{Multivariate Model Sizes}
\label{sec:vector_model_sizes}
\begin{figure}[t]
\centering
\includegraphics[width=0.9\columnwidth]{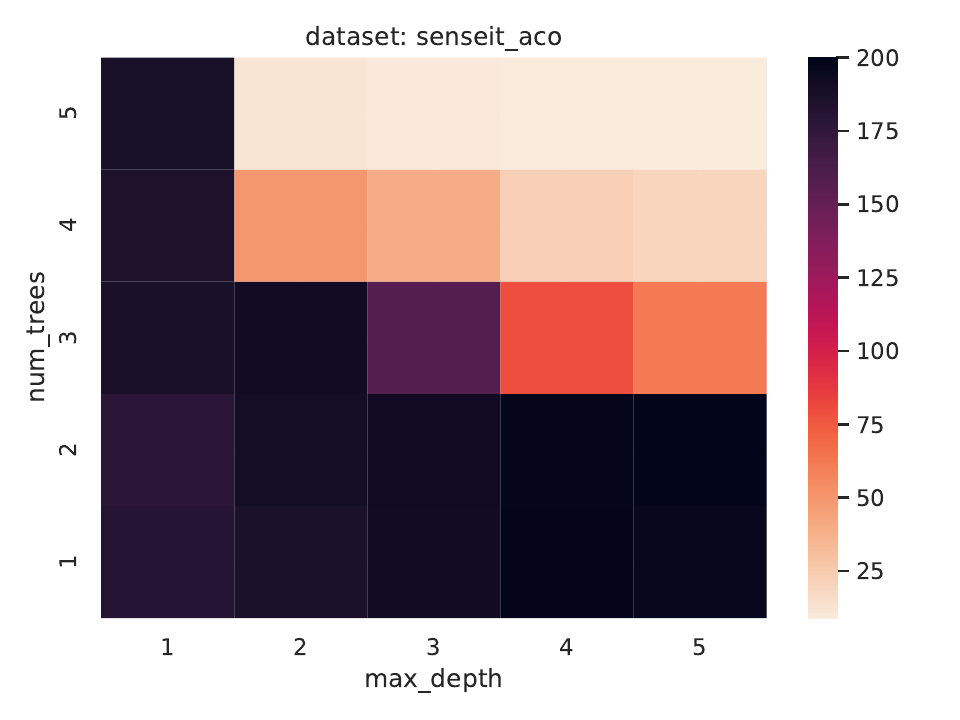}
\caption{Improvements in test $F1$-macro for the dataset \emph{senseit-aco} for different sizes of $GBM$ models. Here, model size is determined by both \emph{max\_depth} per tree and \emph{number of trees}. Greater improvements are seen at lower sizes.}
\label{fig:sample_gbm_result}
\end{figure}

Our technique is applicable even when the model size has more than one attribute. This is because Algorithm \ref{algo:main_with_data} delegates size enforcement to $train_{\mathcal{I}, g}$. Consider GBMs, where we might consider a bivariate size, $\eta= [max\_depth, num\_trees]$; here the quantities respectively denote the maximum depth allowed for each constituent DT in a GBM, and the number of DTs in the GBM. In Figure \ref{fig:sample_gbm_result} we show how improvements for GBMs vary when $1 \leq max\_depth \leq 5 $ ($x$-axis) and $1 \leq num\_trees \leq 5 $ ($y$-axis); the oracle used is a GBM as well (unconstrained in size), and the dataset used is \emph{senseit-aco}.The improvements are averaged over three runs. We observe the familiar pattern that as model sizes increase, in terms of both $max\_depth$ and $num\_trees$, improvements decrease. Results over more datasets is visualized in \S \ref{sec:vector_model_sizes_appendix}.

\section{Different Feature Spaces}
\label{sec:different_feature_space}
So far we have assumed identical feature spaces for the interpretable and  oracle models. But this is not necessary since all that is required of the oracle are uncertainty scores, irrespective of how the oracle arrives at them. 

\begin{figure}[htbp]
\centerline{\includegraphics[width=0.9\columnwidth]{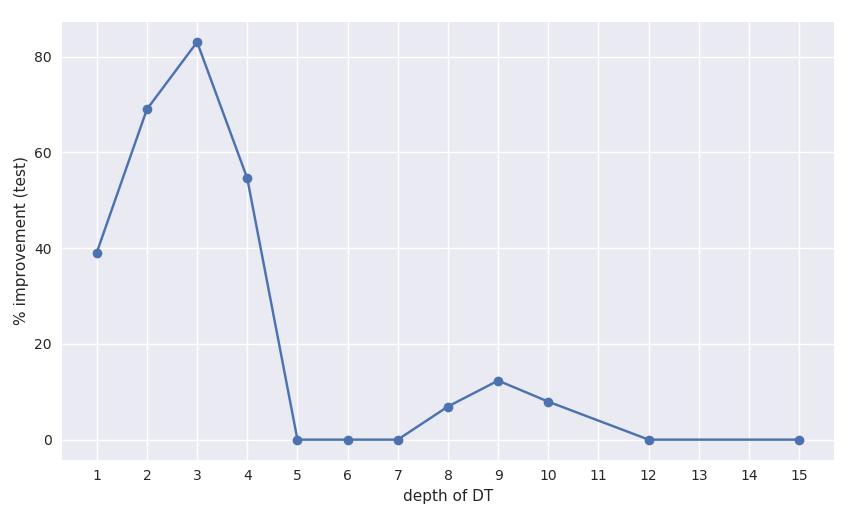}}
\caption{Improvements $\delta F1$ are shown for different depths of the DT. }
\label{fig:rnn_to_dt_improvements_main}
\end{figure}

To highlight this, we consider the classification task of predicting nationalities from surnames \cite{NLP-Pytorch}. This dataset contains $18$ nationalities. We use a \emph{Gated Recurrent Unit (GRU)} \cite{cho-etal-2014-learning}  as our oracle.
This is trained on the surnames presented as character sequences. As the interpretable model, we use a CART DT  that uses character \emph{n-grams} as input. 
The improvements obtained are shown in Figure \ref{fig:rnn_to_dt_improvements_main}.   This is practically powerful since it allows for translating information between models of varied capabilities. See \S \ref{sec:different_feature_space_appendix} for additional details.

\section{Conclusion and Future work}
\label{sec:conclusion}
In this work we presented a iterative model-agnostic technique to obtain good size-accuracy trade-offs. Conveniently, there is only one hyperparameter to set (the number of iterations). Further, it can accommodate multivariate model sizes and can be used with differing feature spaces between the oracle and the interpretable model. We believe that the strategy of manipulating the training distribution to improve accuracy presents multiple interesting possibilities. 

For future work, we think the following ideas are promising: (a) A specialized version for differentiable models/losses. The current technique is attractive because of its broad applicability, but it can be made more efficient by exploiting differentiability (when possible). Techniques such as \emph{bilevel optimization}, e.g.,  \citet{hoag}, might be useful here to learn instance weights. (b) We noted that improvements diminish with increasing model size. An interesting direction to explore is whether this effect might be delayed by applying the technique separately to smaller models obtained from decomposing a larger model, e.g., subtrees within a larger tree.

\clearpage

\bibliography{refs}
 \section{Reproducibility checklist}

This paper:
\begin{itemize}
    \item Includes a conceptual outline and/or pseudocode description of AI methods introduced. Yes.
    \item Clearly delineates statements that are opinions, hypothesis, and speculation from objective facts and results. Yes.
    \item Provides well marked pedagogical references for less-familiar readers to gain background necessary to replicate the paper.  Yes.
\end{itemize}

Does this paper make theoretical contributions? No.

If yes, please complete the list below.
\begin{itemize}
    \item All assumptions and restrictions are stated clearly and formally. NA.
    \item All novel claims are stated formally (e.g., in theorem statements). NA.
    \item Proofs of all novel claims are included. NA.
    \item Proof sketches or intuitions are given for complex and/or novel results. NA.
    \item Appropriate citations to theoretical tools used are given. NA.
    \item All theoretical claims are demonstrated empirically to hold. NA.
    \item All experimental code used to eliminate or disprove claims is included. NA.
\end{itemize}

Does this paper rely on one or more datasets? Yes.

If yes, please complete the list below.
\begin{itemize}
    \item A motivation is given for why the experiments are conducted on the selected datasets. Yes.
    \item All novel datasets introduced in this paper are included in a data appendix. NA.
    \item All novel datasets introduced in this paper will be made publicly available upon publication of the paper with a license that allows free usage for research purposes. NA.
    \item All datasets drawn from the existing literature (potentially including authors’ own previously published work) are accompanied by appropriate citations. Yes.
    \item All datasets drawn from the existing literature (potentially including authors’ own previously published work) are publicly available. Yes.
    \item All datasets that are not publicly available are described in detail, with explanation why publicly available alternatives are not scientifically satisficing. NA.
\end{itemize}

Does this paper include computational experiments? Yes.

If yes, please complete the list below.
    \begin{itemize}
        
    \item Any code required for pre-processing data is included in the appendix. Yes.
    \item All source code required for conducting and analyzing the experiments is included in a code appendix. Yes.
    \item All source code required for conducting and analyzing the experiments will be made publicly available upon publication of the paper with a license that allows free usage for research purposes. Yes.
    \item All source code implementing new methods have comments detailing the implementation, with references to the paper where each step comes from. Yes.
    \item If an algorithm depends on randomness, then the method used for setting seeds is described in a way sufficient to allow replication of results. Yes.
    \item This paper specifies the computing infrastructure used for running experiments (hardware and software), including GPU/CPU models; amount of memory; operating system; names and versions of relevant software libraries and frameworks. Yes.
    \item This paper formally describes evaluation metrics used and explains the motivation for choosing these metrics. Yes.
    \item This paper states the number of algorithm runs used to compute each reported result. Yes.
    \item Analysis of experiments goes beyond single-dimensional summaries of performance (e.g., average; median) to include measures of variation, confidence, or other distributional information. Yes.
    \item The significance of any improvement or decrease in performance is judged using appropriate statistical tests (e.g., Wilcoxon signed-rank). Yes.
    \item This paper lists all final (hyper-)parameters used for each model/algorithm in the paper’s experiments. Yes.
    \item This paper states the number and range of values tried per (hyper-) parameter during development of the paper, along with the criterion used for selecting the final parameter setting. Yes.
\end{itemize}

\clearpage
\appendix
\section{Appendix}
\subsection{Uncertainty Metrics}
\label{sec:appendix_unc_metrics}

Some other popular uncertainty metrics are:

\begin{enumerate}
    \item \textbf{Least confident}: we calculate the extent of uncertainty w.r.t. the class we are most confident about:
    \begin{equation}
    u_M(x) = 1-\max_{y_i \in  \{1,2,..., C\}  } M(y_i|x)
    \label{eqn:leastconfident}
    \end{equation}
    Here, we have $C$ classes, and $M(y_i|x)$ is the probability score produced by the model\footnote{The possibly confusing name ``least confident'' for this idea originated within the context of uncertainty sampling, where we are interested in sampling the most uncertain point, $x^* = \argmin_x [\max_{y_i \in  \{1,2,..., C\}  } M(y_i|x)]$, which may be considered to be the instance with the ``least most confident label''.}.

    \item \textbf{Entropy}: this is the standard  Shannon entropy measure calculated over class prediction confidences:
    \begin{equation}
        u_M(x) = \sum\limits_{y_i \in  \{1,2,..., C\} } -M(y_i|x)\log M(y_i|x)
    \end{equation}
    
\end{enumerate}

We do not use the \emph{least confident} metric since it completely ignores confidence distribution across labels. While \emph{entropy} is quite popular, and does take into account the confidence distribution, we do not use it since it reaches its maximum for only points for which the classifier must be equally ambiguous about \emph{all} labels; for datasets with many labels (one of our experiments uses a dataset with $26$ labels - see Table \ref{tab:datasets}) we may never reach this maximum. 

\begin{figure}[!h]
\centerline{\includegraphics[width=0.9\columnwidth]{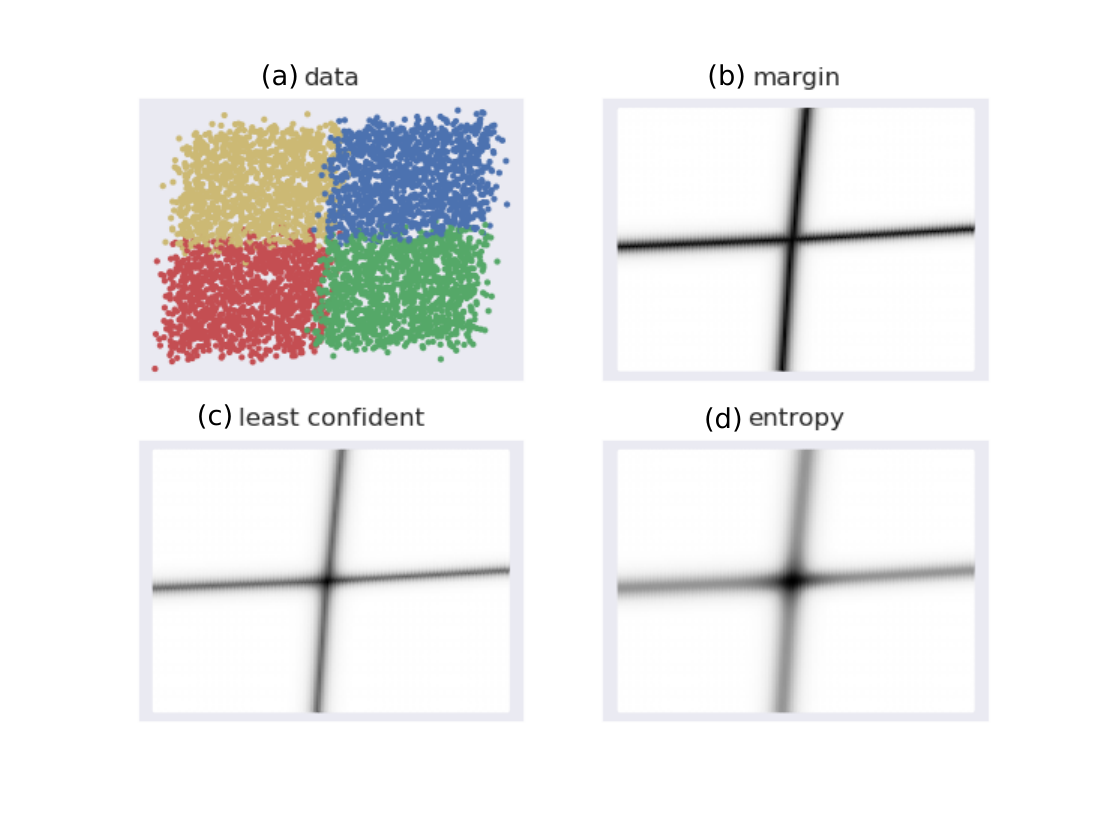}}
\caption{Visualizations of different uncertainty metrics. (a) shows a 4-label dataset on which linear SVM is learned. (b), (c), (d) visualize uncertainty scores based on different metrics, as per the linear SVM, where darker shades imply higher scores.}
\label{fig:uncertainty_types}
\end{figure}

Fig \ref{fig:uncertainty_types} visually shows what uncertainty values look like for the different metrics. Panel (a) displays a dataset with 4 labels. A probabilistic \emph{linear Support Vector Machine (SVM)} is learned on this, and uncertainty scores corresponding to the metrics ``margin'', ``least confident'' and ``entropy'' are visualized in panels (b), (c) and (d) respectively. Darker shades of gray correspond to high uncertainty. Observe that only the ``margin'' metric in panel (b) achieves scores close to $1$ at the two-label boundaries.

There is no best uncertainty metric in general, and the choice is usually application specific \citep{settles2009active}.

\subsection{Sampling from the IBMM}
\label{sec:sampling_ibmm}

Given our representation, the procedure to sample $N_s$ points, from a dataset $D$, using an oracle $M_O$ is shown in Algorithm \ref{alg:sample_oracle}. We also explain the steps below:
 \begin{enumerate}
    \item Determine partitioning over the $N_s$ points induced by the $DP$. We use the \emph{Chinese Restaurant Process} \cite{ref10.1007/BFb0099421} for this.
    Let's assume this step produces $k$ partitions $\{c_1, c_2, ..., c_k\}$ and quantities $n_i \in \mathbb{N}$ where $\sum_{i=1}^k n_i=N$. Here, $n_i$ denotes the number of points that belong to partition $c_i$.
    \item We determine the $Beta(A_i, B_i)$ component for each $c_i$ by sampling from the priors, i.e.,   $A_{i} \sim scale \times Beta(a, b)$ and $B_{i} \sim scale \times Beta(a', b')$.
    
    \item Repeat for each $c_i$: for each instance-label pair $(x_j, y_j)$ in our training dataset, we calculate the oracle uncertainty score, $u_{M_O}(x_j)$. We then calculate $p_j = c \cdot Beta(u_{M_O}(x_j)|A_i, B_i)$. $c$ is a normalizing constant that scales the probabilities across instances to sum to 1. The quantities $p_j$ are used as sampling probabilities for various $(x_j, y_j)$, and $n_i$ points are sampled with replacement based on them. 
\end{enumerate}

\begin{algorithm}[tb]
 \KwData{Sample size $N_s$, oracle $M_O$, dataset  $D=\{(x_i, y_i)\}_{i=1}^N$, IBMM parameters $\Psi=\{\alpha, a, b, a', b'\}$}
 \KwResult{Sample $D'$, where $|D'|=N_s$}
$D'=\{\}$ \tcp{assumed to be a multiset}
$\{(c_1, n_1), (c_2, n_2), ..., (c_k, n_k)\} \gets $ partition $N_s$ using the $DP$ \tcp{Here $\sum_{i=1}^k n_i=N_s$.}

\For {$i\gets1$ \KwTo $k$}{
$A_{i} \sim scale \times Beta(a, b)$\\
$B_{i} \sim scale \times  Beta(a', b')$ \\
\For {$j\gets1$ \KwTo $N$}{
    $p_j \gets c \cdot Beta(u_{M_O}(x_j);A_i, B_i)$ \tcp{$c$ is a normalizing constant s.t. $\sum_i^N c \cdot p_j = 1$.}
}
$temp \gets$ sample with replacement $n_i$ instance-label pairs based on $p_j$ \\
$D' \gets D' \uplus temp$ \tcp{$\uplus$ is the multiset sum}
 }

 \Return $D'$
 \caption{Sample based on uncertainties and $\Psi$}
 \label{alg:sample_oracle}
\end{algorithm}

\subsection{Default Parameters}
\label{sec:default_params}

    The optimizer we use, TPE, requires \emph{box constraints}. Here we specify our search space for the optimization variables, $\Phi$ in Algorithm \ref{algo:main_with_data}:
    \begin{enumerate}
        \item $p_o$: We want to allow the algorithm to pick an arbitrary fraction of samples from the original data; we set $p_o \in [0, 1]$.
        \item $N_s$: We set $N_s \in [400, 10000]$. The lower bound ensures we have statistically significant results. The upper bound is set to a reasonably large value.
        
        \item $\{a, b, a', b'\}$: Each of these parameters are allowed a range $[0.1, 10]$ to allow for a wide range of shapes for the component $Beta$ distributions. 
        
        \item $scale$: We fix $scale=10000$ for our experiments, to allow for $A_i$ and $B_i$ to model skewed distributions where shape parameter large values might be required. For small values, the algorithm adapts by learning the appropriate  $\{a, b, a', b'\}$.
        
        \item $\alpha$: For a DP, $\alpha \in \mathbb{R}_{>0}$.
        We use a lower bound of $0.1$. 
        
        To determine the upper bound, we rely on the following empirical relationship \citep{doi:10.1002/sim.2666} between the number of components $k$ and $\alpha$:
        \begin{equation}
        \label{eqn:dp_alpha}
            E[k|\alpha] \approx 5\alpha+2 
        \end{equation}
        
        We empirically estimated a fairly inclusive upper bound on the number of components to be $500$, which provides us the $\alpha$ upper bound of $99.6$. Thus, we use $\alpha \in [0.1, 99.6]$.

\end{enumerate}

\subsection{Notes on the Main Algorithm}
\label{sec:notes_algorithm}
We provide some additional details in reference to the main algorithm - Algorithm \ref{algo:main_with_data} - in the paper. For convenience, we reproduce the algorithm here, as Algorithm \ref{algo:main_with_data_appendix}. Our notes follow:

\begin{algorithm}[tb]
 \KwData{Dataset $D$, model size $\eta$, $train_{\mathcal{O}, h}()$, $train_{\mathcal{I}, g}()$, iterations $T$}
 \KwResult{Optimal parameters $\Phi^*$, test set accuracy $s_{test}$ at $\Phi^*$, and interpretable model $M^*$ at $\Phi^*$}
Create splits $D_{train}, D_{val}, D_{test}$ from $D$, stratified wrt labels. Here $|D_{train}|: |D_{val}|: |D_{test}|:: 60:20:20$.\\
$M_O \gets train_{\mathcal{O},h}(D_{train}, *)$ \\

\For {$t\gets1$ \KwTo $T$ } { 
   $\Phi_{t} \gets suggest(s_0, s_1, ... s_{t-1}, \Phi_0, \Phi_1, ..., \Phi_{t-1})$ \tcp{$s_0, \Phi_{0}$ initialized at $t=0$, see text. Note: $\Phi_t = \{\Psi_t, N_{s,t}, p_{o,t}\}$ where $\Psi_t = \{\alpha_t, a_t, b_t, a'_t, b'_t\}$.}

$N_o \gets p_{o,t} \times N_{s,t}$  \\
$N_u \gets N_{s\_t} - N_o$ \\
$D_o \gets$ uniformly sample with replacement  $N_o$ points from  $D_{train}$\\ 
$D_u \gets$ sample $N_u$ points from $D_{train}$  using the DP-based IBMM given current values for $N_u, M_O, D_{train}, \Psi_t$ \tcp{see Algorithm \ref{sec:sampling_ibmm} for details}
  $D_s \gets D_o \uplus D_u$ \tcp{$D_o$, $D_u$ are assumed to be multisets}
  $M_t \gets train_{\mathcal{I}, g}(D_s, \eta) $\\
  $s_t \gets acc(M_t, D_{val})$
 }
$t^* \gets \argmax_t{\{s_1, s_2, ..., s_{T-1}, s_T\}}$\\
$\Phi^* \gets \Phi_{t^*}$\\
$M^* \gets M_{t^*}$\\
$s_{test} \gets acc(M^*, D_{test})$\\
\Return $\Phi^*$, $s_{test}$, $M^*$
 \caption{Learning interpretable model using oracle - reproduction of Algorithm \ref{algo:main_with_data}.}
 \label{algo:main_with_data_appendix}
\end{algorithm}

\begin{enumerate}
    \item We will consider the initialization to happen at $t=0$, while the iterations range from $1$ to $T$. $\Phi_0$ is set to: $\alpha=0.1, a=1,b=1,a'=1,b'=1, N_s=|D_{train}|, p_o=1$. A model is constructed based on $\Phi_0$ and a score $s_0$ is recorded. $(\Phi_0, s_0)$ serve as the history for the iteration at $t=1$. The values for $\alpha, a, b, a', b'$ carry no significance and are arbitrary, since setting $p_o \to 1$ forces sampling only from the original distribution. Combined with $N_s=|D_{train}|$, this setting mimics the baseline, i.e., training the interpretable model without our algorithm, thus providing the optimizer with a good initial reference point in its search space.

    \item The optimizer is represented by the function call $suggest()$ which takes as input all past parameter values and validation scores. $suggest()$ denotes a generic optimizer; not all optimizers require this extent of historical information.
    \item While the training algorithm for the oracle, $train_{\mathcal{O}, h}()$ is taken as input, a pre-constructed oracle $M_O$ may also be used. This would eliminate the oracle training step in line 2. 
    \item $acc()$ on the validation data, $D_{val}$, serves as both the objective and fitness function.
    \item Evaluation on the test set, $D_{test}$ is done only once, in line 16, with the model that produces the best validation score.
    \item Since we sample with replacement, both temporary datasets $D_o$ and $D_u$, procured from uniformly sampling the original training data and sampling based on uncertainties respectively, are multisets. Accordingly, line 9 uses the multiset sum operator $\uplus$ to combine them.
    \item $M_t$ is created (line 10) with limited or no hyperparameter search using simple random validation, i.e., a stratified (by labels) random sample of size $0.2N_{s,t}$ is used as the validation set. A restricted search is performed because often hyperparameters are correlated with model size, and setting them to particular values would fail to produce a model of the required size $\eta$. As an example, consider DTs: setting a high threshold for the number of instances in a node for it be split (hyperparameter \emph{min\_samples\_split} in \emph{scikit-learn's} \citep{scikit-learn} implementation) would produce only short trees. 

    We don't use cross-validation since at small values of $N_{s,t}$, the amount of training data, i.e.,  $(\frac{k-1}{k}) N_{s,t}$ for $k$-folds,  may become too small to obtain a good model. For example, for 3-folds, the training data size is $0.67 N_{s,t}$. The data shortage problem can be addressed by increasing the number of folds, but that also increases the running time per iteration owing to the larger number of models that now need to be trained. As a practical compromise, we perform simple validation \emph{thrice} and average the outcomes. This number is configurable, and may be decreased for models that are expensive to train.

    \item Since the validation score $s_t$ (line 11) needs to be reliable, in our implementation we repeat lines 7-10 \emph{thrice} and use the averaged validation score as $s_t$.
    \item Class imbalance is accounted for in our implementation when training model $M_t$ in line 10. We either balance the data by sampling (this is the case with a \emph{Linear Probability Model}), or an appropriate cost function is used to simulate balanced classes (this is the case with DTs and GBMs).

\end{enumerate}

    It is important to note here that \emph{$D_{val}$ and $D_{test}$ are not modified by our algorithm in any way}, and therefore $s_t$ and $s_{test}$ measure the accuracy on the original distribution. 

\subsection{Smoothing the Optimization Landscape}
\label{ssec:smoothe_opt}
A practical consideration in our implementation is if we might facilitate finding the maxima $\Phi^*$ in Algorithm \ref{algo:main_with_data}?

Since BayesOpt algorithms model the response surface of the actual objective function using a finite number of evaluations ($s_t$ in Algorithm \ref{algo:main_with_data}), a certain degree of \emph{smoothness} is assumed \citep{7352306,Brochu2010ATO}.
Here, the optimization variables $\Phi$ influence the objective value $s$ via this indirect chain: $\Phi_t \rightarrow D_s \rightarrow M_t \rightarrow s_t$ (symbols as in Algorithm \ref{algo:main_with_data}), and for BayesOpt to work well, it is required that small changes in $\Phi_t$ result in small changes in $s_t$.

\begin{figure}[htbp]
\includegraphics[width=0.95\columnwidth]{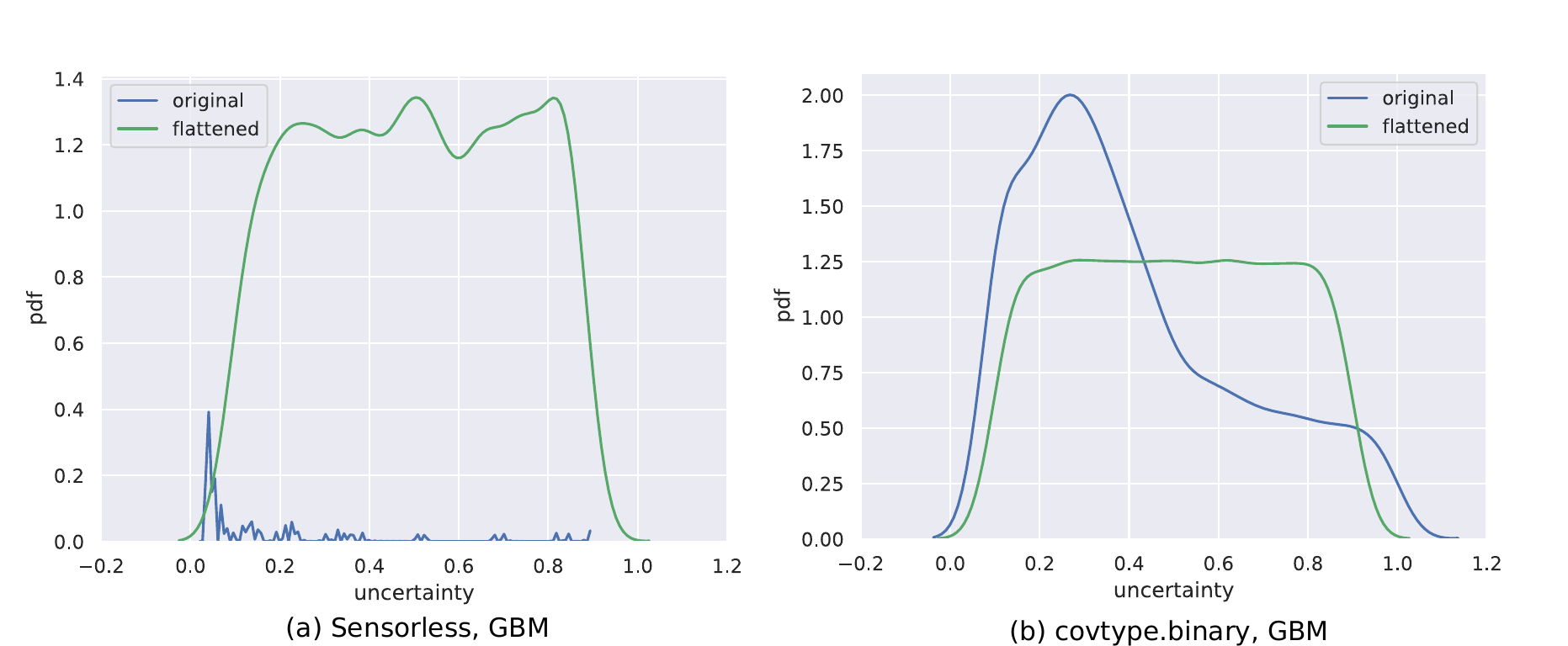}
\caption{Example of curve-flattening, for datasets (a) \texttt{Sensorless} and (b) \texttt{covtype.binary}. The uncertainty scores shown are obtained using the $GBM$ oracle.}
\label{fig:flattening}
\end{figure}

However, we have noticed that an oracle might produce uncertainty score distributions that are ``spiky'' or ``jagged'' - as an example,  see the curve labelled ``original'' in Figure \ref{fig:flattening}(a); which leads us to hypothesize that this principle is violated in general. A spiky distribution implies that small shifts $\Phi_t + \Delta \Phi_t$ may lead to sampling of instances with very different uncertainties; and since such instances may occur in regions far from those indicated by $\Phi_t$, they produce models with different class prediction behavior. This indirectly causes a disproportionate shift in $s_t$. While, in theory, a good BayesOpt algorithm should adapt to such problem characteristics, in practice they make the optimization problem harder, especially when the optimization budget is small.

To address this, we ``flatten'' the distribution\footnote{Distribution transformations have a long history in statistics, e.g., \emph{power transforms} like the \emph{Box-Cox} \citep{boxcox10.2307/2984418} and \emph{Yeo-Johnson} \citep{yeo10.2307/2673623} transforms. Within ML, \emph{Batch Normalization} \citep{BN10.5555/3045118.3045167} is a popular example of a distribution transformation applied to a loss landscape \citep{how_batchnorm_works}.} within $[0, 1]$. 
Our transformation is simple: we divide the interval $[0,1]$ into $B$ bins, and map approximately $ |D_{train}|/B $ uncertainty scores to each bin, while maintaining order between the original and mapped scores. Within a bin, the mapped scores are linearly spread across its range. This distributes the mapped scores approximately uniformly in the range $[0, 1]$. The algorithm is detailed  in Algorithm \ref{alg:binning}.

\begin{algorithm}[tbh]
 \KwData{$\{u(x_1), u(x_2), ..., u(x_N)\}, \text{ number of bins }B$}
 \KwResult{$\{u'(x_1), u'(x_2), ..., u'(x_N)\}$ }
 $bin\_size \gets \lceil N/B \rceil, bin\_range \gets  1/B $ \\
 $bin\_min \gets [\;], bin\_max \gets [\;]$ \\
 Let $sortedIndex(i) \in \{1, 2,...,N\}$ be the index of $u(x_i)$ in the sequence of scores ordered by non-decreasing values. \\

\For {$j\gets1$ \KwTo $B$}{
    $bin\_min[j] \gets \min \{u(x_i)|i \in \{1, 2,...,N\}\land sortedIndex(i)=j\}$ \\
    $bin\_max[j] \gets \max \{u(x_i)|i \in \{1, 2,...,N\} \land sortedIndex(i)=j\}$ \\
}

\For {$i\gets1$ \KwTo $N$}{
    $j \gets sortedIndex(i)$\\
    $bin\_num \gets \lceil j/bin\_size \rceil$ \\
    
    $boundary\_low \gets (bin\_num - 1) \times bin\_range +\delta$\\
    $boundary\_high \gets bin\_num \times bin\_range - \delta$\\
    $u'(x_i) \gets low + \frac{u(x_i)- bin\_min[j]}{bin\_max[j]-bin\_min[j]}\times(boundary\_high-boundary\_low)$ \\
}
\Return $\{u'(x_1), u'(x_2), ..., u'(x_N)\}$
 \caption{Flatten distribution of uncertainty scores $\{u(x_1), u(x_2), ..., u(x_N)\}$}
 \label{alg:binning}
\end{algorithm}

Figure \ref{fig:flattening} visualizes the process of flattening. The original and modified uncertainty distributions for the datasets \texttt{Sensorless} and \texttt{covtype.binary} are shown in Figure \ref{fig:flattening}(a) and \ref{fig:flattening}(b) respectively.

 While \texttt{Sensorless} appears to have a non-smooth distribution, and flattening here might help, this seems redundant for \texttt{covtype.binary}. \emph{However, since this step is computationally cheap, we perform this for all our experiments, saving us the effort of  assessing its need.} 

Our transformation is invertible, which is useful in analyzing the observations from our experiments. Note however, it is not differentiable because of the discontinuities at the bin-boundaries; we also don't require this property.

The transformation affects line 7 in Algorithm \ref{alg:sample_oracle}. Instead of sampling based on the actual oracle uncertainty scores:

\begin{equation}
   p_j \gets Beta(u_{M_O}(x_j);A_i, B_i)
\end{equation}
we sample based on the transformed uncertainty scores, $u'_{M_O}(x_j)$:

\begin{equation}
   p_j \gets Beta(u'_{M_O}(x_j);A_i, B_i)
\end{equation}
In \S \ref{sec:effect_smoothing} we show that smoothing indeed has a positive effect.

\subsection{Effect of Smoothing}
\label{sec:effect_smoothing}

We first consider the question: does flattening (\S \ref{ssec:smoothe_opt}) help?  Table \ref{tab:effect_of_flattening} contrasts \emph{improved} $F1$ scores obtained without (rows denoted as ``original'') and with (denoted ``flattened'') flattening  the uncertainty distribution. This is shown for the datasets \texttt{Sensorless} and \texttt{covtype.binary}, for model $size \in \{1,2,3\}$, with $model=LPM$ and $ oracle=GBM$. Two different parameter settings are used: (a) In Setting 1, maximum allowed $Beta$ components are $500$ and $scale=10000$ (b) Setting 2 looks at much lower values of these parameters where maximum allowed components is $50$ and $scale=10$. The scores presented are the average over three trials.

\begin{table}\scriptsize
\centering
\caption{Improved scores averaged over three trials, shown for different parameter settings, with and without flattening. Here, Setting 1 is $\{max\_components=500, scale=10000\}$ and Setting 2 is $\{max\_components=50, scale=10\}$. ``curr.'' signifies this is the current setting for our experiments in the main paper, while ``low'' signifies lower values of parameters. Highlighted cells indicate positive effect of flattening.} 
\label{tab:effect_of_flattening}
\hspace{0.5pt}
\resizebox{0.95\columnwidth}{!}{\begin{tabular}{llrrrrrr}
\toprule
\multicolumn{1}{r}{}& \multicolumn{1}{r}{}& \multicolumn{3}{c}{Setting 1 (curr.)} & \multicolumn{3}{c}{Setting 2 (low)} \\ 
\cmidrule(lr){3-5}
\cmidrule{6-8}
dataset & dist. & $1$ & $2$  & $3$ & $1$ & $2$  & $3$ \\
\toprule
Sensorless & original & \cellcolor{blue!25} 0.39 & 0.54 & 0.57 & \cellcolor{blue!25} 0.38 &\cellcolor{blue!25}  0.42 & \cellcolor{blue!25} 0.41 \\ 
 & flattened & \cellcolor{blue!25} 0.44 & 0.53 & 0.55 & \cellcolor{blue!25} 0.43 & \cellcolor{blue!25} 0.54 & \cellcolor{blue!25} 0.59 \\ 
 covtype.binary & original & 0.66 & \cellcolor{blue!25} 0.69 & 0.71 & 0.64 & \cellcolor{blue!25} 0.66 & 0.71 \\ 
 & flattened & 0.68 &\cellcolor{blue!25} 0.73 &0.73 & 0.65 &  \cellcolor{blue!25}0.71 & 0.71 \\ 
\bottomrule
\end{tabular}
}
\end{table}

We observe that while flattening influences results, other parameters determine the magnitude of its effect. At Setting 1, \texttt{Sensorless} is affected at $size=1$ (flattening is better), but at higher sizes the differences seem to be from random variations across trials. At Setting 2 however, the differences are seen for $size \in \{1,2,3\}$  (flattening is better). For \texttt{covtype.binary} only $size=2$ seems to be affected in either setting.

Recall we had noted in Figure \ref{fig:flattening} that the datasets \texttt{Sensorless} and \texttt{covtype.binary} have  non-smooth and smooth uncertainty distributions respectively. The observations in Table \ref{tab:effect_of_flattening} align well with the expectation that \texttt{Sensorless} is positively affected by the transformation, while results for \texttt{covtype.binary} remain mostly unchanged.

Based on these tests, we hypothesize that for non-smooth uncertainty distributions, flattening makes our technique robust across parameter settings. It does not affect smooth distributions in a significant way. Of course, rigorous and extensive tests are required to conclusively establish this effect.

\subsection{Datasets}
\label{sec:datsets}
Table \ref{tab:datasets} provides details about the various datasets used in the experiments in \S \ref{sec:experiments}. All of these are publicly available on the \emph{LIBSVM} website \protect\citep{CC01a}. 

The ``Label Entropy'' column indicates how balanced a dataset  is wrt its classes. For a dataset with $N$ instances and $C$ labels, this is calculated as:
\begin{align}
    \text{Label Entropy} &= \sum_{j \in \{1,2,..., C\}} - p_j \log_C p_j\\
    \text{Here, }\; p_j &= \frac{|\{x_i|y_i=j\}|}{N} \nonumber
\end{align}
$\text{Label Entropy} \in [0,1]$, where values close to $1$ denote the dataset is nearly balanced, and values close to $0$ represent relative imbalance.

\begin{table*} \small
\centering
\caption{We use the following datasets available on the LIBSVM website \protect\citep{CC01a}. Their original source is mentioned in the ``Description'' column. $10000$ instances from each dataset are used. 
A $train:val:test$ split ratio of $60:20:20$ is used for $D_{train}, D_{val}$ and $D_{test}$ in Algorithm \ref{algo:main_with_data}. The splits are stratified wrt labels.}\label{tab:datasets}
\vspace{0.75pt}
\hspace{0.75pt}
\resizebox{1\textwidth}{!}{
\begin{tabular}{llrrrp{7cm}}
\toprule
S.No. & Dataset & Dimensions & \# Classes & Label Entropy & Description\\ 
\midrule
1& cod-rna & 8 & 2 & 0.92 & Predict presence of non-coding RNA common to a pair of RNA sequences, based on individual sequence properties and their similarity \citep{Uzilov2006}.\\[0.3cm]
2& ijcnn1 &  22 &  2 & 0.46 & Time series data produced by an internal combustion engine is used to predict normal engine firings vs misfirings \citep{ijcnn1_data}. Transformations as in \citet{Chang01ijcnn2001}.\\[0.3cm]
3& higgs &  28 &  2 & 1.00 & Predict if a particle collision produces Higgs bosons or not, based on collision properties \citep{Baldi2014}.\\[0.3cm]
4& covtype.binary &  54 &  2 & 1.00 & Modification of the \emph{covtype} dataset (see row 12), where classes are divided into two groups \citep{NIPS2001_1949}.\\[0.3cm]
5& phishing &  68 &  2 & 0.99 &Various website features are used to predict if the website is a \emph{phishing} website \citep{phishing_original}. Transformations used as in \citet{phishing_data_transform}\\[0.3cm]
6& a1a &  123 &  2 & 0.80 &Predict whether a person makes over 50K a year, based on census data variables \citep{Dua:2019}. Transformations as in \cite{platt1998fast}.\\[0.3cm]
7& pendigits &  16 &  10 & 1.00 & Classify handwritten digit samples into the digits 0-9 \citep{Alimoglu96methodsof,Dua:2019}.\\[0.3cm]
8& letter &  16 &  26 & 1.00 & Images of the capital letters A-Z were produced by random distortion of these characters from 20 fonts. The task is to classify these character images as one of the original letters \citep{10.5555/212782}. Transformations as in \cite{multiclass_svm_compare}.\\[0.3cm]
9& Sensorless &  48 &  11 & 1.00 & Based on phase current measurements of an electric motor, predict different error conditions \citep{sensorless_paper}. We use the transformations from \cite{10.1162/neco_a_01088}. \\[0.3cm]
10& senseit\_aco &  50 &  3 & 0.95 & Predict vehicle type using acoustic data gathered by a sensor network \citep{10.1016/j.jpdc.2004.03.020}.\\[0.3cm]
11& senseit\_sei &  50 &  3 & 0.94 & Predict vehicle type using seismic data gathered by a sensor network \citep{10.1016/j.jpdc.2004.03.020}.\\[0.3cm]
12& covtype &  54 &  7 & 0.62 & Predict forest cover type from cartographic variables \citep{Dean1998ComparisonON,Dua:2019}.\\[0.3cm]
13& connect-4 &  126 &  3 & 0.77 & Predict if the first player wins, loses or draws, based on board positions of the board game \emph{Connect Four} \citep{Dua:2019}.\\
\bottomrule
\end{tabular}
}\end{table*}

\subsection{Validation Results}
\label{sec:appendix_val_results}
An extended version of the results shown in Table \ref{tab:internal_improvements_short} are presented here in Table \ref{tab:internal_improvements}. This shows results for all combinations of models and oracles: $\{LPM, DT\} \times \{GBM, RF\}$.

\begin{table*}[hbt!] \small
            \centering
            \caption{\small This table shows the average improvements, $\delta F1$, over five runs for different combinations of models and oracles: $\{LPM, DT\} \times \{GBM, RF\}$. This is an extended version of the results in Table \ref{tab:internal_improvements_short}. The improvements are measured relative to the model at the first iteration.The best improvement for a model size and oracle is indicated in bold. Here, $\delta F1 \in (-\infty, \infty)$. Negative improvements are shown in \unexpcol{underlined}.}\label{tab:internal_improvements}
            \hspace{0.5pt}
            \resizebox{\textwidth}{!}{\begin{tabular}{lllllllllllllllll}
            \toprule [1.1pt]
dataset & model\_ora & 1 & 2 & 3 & 4 & 5 & 6 & 7 & 8 & 9 & 10 & 11 & 12 & 13 & 14 & 15\\
 \toprule [1.1pt]
cod-rna & lpm\_gbm & 1.39 & 12.53 & \textbf{14.76} & \textbf{15.73} & 14.97 & 12.00 & 0.00 & \textbf{0.08} & - & - & - & - & - & - & - \\  
 & lpm\_rf & \textbf{2.66} & \textbf{13.91} & 14.69 & 15.34 & \textbf{16.06} & \textbf{12.49} & \textbf{8.30} & 0.00 & - & - & - & - & - & - & - \\ [3pt] 
 & dt\_gbm & \textbf{0.00} & \textbf{0.00} & 0.00 & 1.26 & 0.00 & \textbf{0.00} & 0.00 & 0.00 & \unexpcol{-0.28} & 0.08 & - & - & - & - & - \\  
 & dt\_rf & 0.00 & 0.00 & \textbf{1.78} & \textbf{2.28} & \textbf{0.39} & \unexpcol{-0.02} & \textbf{0.17} & \textbf{0.47} & \textbf{0.00} & \textbf{0.72} & - & - & - & - & - \\  
\midrule
ijcnn1 & lpm\_gbm & \unexpcol{-0.16} & \textbf{3.36} & \textbf{3.93} & 0.00 & \textbf{5.19} & \textbf{4.18} & \textbf{3.85} & \textbf{3.79} & \textbf{3.69} & 2.99 & \textbf{2.97} & 3.21 & 3.11 & 3.26 & 3.02 \\  
 & lpm\_rf & \textbf{0.19} & 2.80 & 3.36 & \textbf{3.65} & 3.33 & 1.94 & 3.58 & 3.30 & 3.46 & \textbf{3.81} & 2.66 & \textbf{4.65} & \textbf{3.99} & \textbf{3.82} & \textbf{4.85} \\ [3pt] 
 & dt\_gbm & 1.96 & 12.00 & \textbf{10.15} & \textbf{11.37} & \textbf{10.63} & 7.18 & 3.63 & \textbf{4.52} & \textbf{2.91} & \textbf{1.78} & \textbf{1.93} & \textbf{2.29} & 1.47 & \textbf{2.26} & 0.00 \\  
 & dt\_rf & \textbf{4.06} & \textbf{12.10} & 8.95 & 10.75 & 10.13 & \textbf{8.25} & \textbf{5.38} & 2.46 & 2.63 & 1.25 & 1.46 & 1.37 & \textbf{1.91} & 0.00 & \textbf{1.38} \\  
\midrule
higgs & lpm\_gbm & \textbf{29.29} & \textbf{17.80} & 11.40 & 6.56 & 3.06 & 2.68 & \textbf{3.16} & \textbf{2.90} & \textbf{2.67} & \textbf{2.82} & 2.65 & 1.79 & 2.62 & 2.19 & 1.63 \\  
 & lpm\_rf & 26.71 & 17.29 & \textbf{15.06} & \textbf{10.60} & \textbf{5.35} & \textbf{4.04} & 2.35 & 2.03 & 1.66 & 1.89 & \textbf{2.91} & \textbf{2.94} & \textbf{3.31} & \textbf{2.58} & \textbf{2.22} \\ [3pt] 
 & dt\_gbm & 0.00 & 0.00 & \textbf{1.86} & 0.26 & 0.93 & 0.45 & - & - & - & - & - & - & - & - & - \\  
 & dt\_rf & \textbf{4.04} & \textbf{1.26} & 1.74 & \textbf{1.32} & \textbf{1.54} & \textbf{0.91} & - & - & - & - & - & - & - & - & - \\  
\midrule
covtype.binary & lpm\_gbm & 76.52 & \textbf{66.39} & \textbf{29.17} & \textbf{12.51} & \textbf{9.18} & \textbf{5.28} & \textbf{4.94} & \textbf{4.56} & \textbf{3.92} & \textbf{3.56} & \textbf{3.62} & \textbf{3.31} & \textbf{2.59} & \textbf{2.83} & \textbf{2.39} \\  
 & lpm\_rf & \textbf{96.77} & 63.38 & 14.36 & 9.61 & 6.79 & 3.94 & 2.93 & 2.81 & 2.96 & 2.84 & 2.31 & 2.26 & 2.00 & 2.43 & 2.22 \\ [3pt] 
 & dt\_gbm & \textbf{0.00} & \textbf{0.00} & \textbf{2.35} & 1.27 & 1.18 & 1.11 & 0.00 & 0.00 & 0.00 & - & - & - & - & - & - \\  
 & dt\_rf & 0.00 & 0.00 & 2.10 & \textbf{2.33} & \textbf{2.44} & \textbf{2.39} & \textbf{1.84} & \textbf{2.19} & \textbf{1.65} & 0.70 & - & 0.89 & - & - & - \\  
\midrule
phishing & lpm\_gbm & \textbf{0.00} & 1.88 & 2.88 & 3.05 & 3.22 & 3.25 & 2.99 & 1.69 & 1.42 & \textbf{1.45} & \textbf{1.29} & 0.00 & 0.00 & 0.00 & 0.00 \\  
 & lpm\_rf & 0.00 & \textbf{2.14} & \textbf{3.29} & \textbf{3.22} & \textbf{3.59} & \textbf{3.79} & \textbf{3.29} & \textbf{2.05} & \textbf{1.42} & 1.44 & 1.24 & \textbf{1.23} & \textbf{1.16} & \textbf{1.26} & \textbf{1.02} \\ [3pt] 
 & dt\_gbm & \textbf{0.00} & 0.00 & \textbf{0.00} & 0.07 & \textbf{0.39} & \textbf{0.00} & \textbf{0.28} & 0.22 & \textbf{0.44} & \textbf{0.23} & 0.00 & \textbf{0.00} & \textbf{0.00} & 0.00 & \textbf{0.00} \\  
 & dt\_rf & 0.00 & \textbf{0.72} & 0.00 & \textbf{0.57} & 0.00 & \unexpcol{-0.17} & 0.13 & \textbf{0.48} & 0.13 & 0.05 & \textbf{0.03} & \unexpcol{-0.03} & \unexpcol{-0.28} & \textbf{0.00} & \unexpcol{-0.16} \\  
\midrule
a1a & lpm\_gbm & \textbf{0.00} & 2.55 & 7.58 & 8.98 & 8.40 & 8.03 & 8.90 & 8.23 & 8.17 & 7.90 & 5.96 & 7.10 & 6.97 & 6.18 & 5.73 \\  
 & lpm\_rf & 0.00 & \textbf{4.17} & \textbf{8.81} & \textbf{9.92} & \textbf{9.88} & \textbf{9.47} & \textbf{8.99} & \textbf{9.31} & \textbf{9.19} & \textbf{9.26} & \textbf{9.33} & \textbf{8.25} & \textbf{7.15} & \textbf{7.55} & \textbf{7.98} \\ [3pt] 
 & dt\_gbm & \textbf{0.00} & 5.54 & 2.39 & 3.84 & \textbf{3.55} & 2.55 & 1.51 & 2.25 & 4.87 & - & - & - & - & - & - \\  
 & dt\_rf & 0.00 & \textbf{6.44} & \textbf{3.36} & \textbf{5.60} & 3.40 & \textbf{5.94} & \textbf{6.06} & \textbf{4.97} & \textbf{4.89} & 4.01 & 4.73 & 5.21 & - & - & 4.53 \\  
\midrule [1.1pt]
pendigits & lpm\_gbm & \textbf{51.39} & \textbf{23.44} & 16.18 & \textbf{8.95} & \textbf{8.84} & \textbf{6.63} & 4.86 & \textbf{1.83} & 2.27 & 2.16 & \textbf{2.44} & 2.16 & \textbf{3.33} & 2.97 & \textbf{2.73} \\  
 & lpm\_rf & 46.28 & 22.74 & \textbf{21.72} & 8.80 & 8.47 & 6.29 & \textbf{6.48} & 1.69 & \textbf{3.03} & \textbf{2.79} & 2.34 & \textbf{2.68} & 2.70 & \textbf{3.02} & 0.00 \\ [3pt] 
 & dt\_gbm & 14.02 & \textbf{6.72} & 5.11 & 13.14 & 6.42 & 4.20 & 2.46 & \textbf{1.09} & \textbf{0.98} & \textbf{0.16} & \unexpcol{-0.26} & \textbf{0.00} & \textbf{0.00} & \textbf{0.00} & \textbf{0.00} \\  
 & dt\_rf & \textbf{21.46} & 4.18 & \textbf{5.22} & \textbf{14.51} & \textbf{7.36} & \textbf{4.55} & \textbf{2.86} & 0.00 & 0.00 & 0.00 & \textbf{0.00} & 0.00 & 0.00 & 0.00 & 0.00 \\  
\midrule
letter & lpm\_gbm & 57.06 & 48.48 & 59.85 & \textbf{29.76} & \textbf{36.09} & 19.27 & 20.37 & 16.08 & 17.55 & 15.16 & 17.26 & 16.51 & 18.46 & 17.19 & 15.55 \\  
 & lpm\_rf & \textbf{61.06} & \textbf{65.34} & \textbf{64.26} & 23.69 & 35.20 & \textbf{26.15} & \textbf{22.10} & \textbf{20.74} & \textbf{20.91} & \textbf{20.31} & \textbf{19.28} & \textbf{21.40} & \textbf{20.77} & \textbf{19.39} & \textbf{18.18} \\ [3pt] 
 & dt\_gbm & \textbf{0.00} & \textbf{13.98} & 25.05 & \textbf{33.96} & 32.05 & 15.49 & \textbf{11.17} & \textbf{0.00} & \textbf{4.26} & \textbf{3.50} & \textbf{1.99} & 0.00 & 0.00 & 0.00 & \textbf{0.00} \\  
 & dt\_rf & 0.00 & 12.21 & \textbf{28.67} & 33.47 & \textbf{33.51} & \textbf{18.41} & 8.10 & 0.00 & 1.84 & 1.21 & 1.31 & \textbf{0.67} & \textbf{0.61} & \textbf{0.11} & \unexpcol{-0.08} \\  
\midrule
Sensorless & lpm\_gbm & 216.47 & \textbf{257.56} & \textbf{178.31} & \textbf{117.01} & \textbf{90.70} & \textbf{83.90} & \textbf{73.50} & \textbf{65.95} & 61.57 & 57.97 & 56.54 & 57.15 & 55.45 & 66.24 & 68.24 \\  
 & lpm\_rf & \textbf{224.18} & 210.28 & 134.44 & 115.00 & 85.85 & 74.96 & 66.77 & 61.10 & \textbf{66.88} & \textbf{64.65} & \textbf{69.00} & \textbf{70.09} & \textbf{72.91} & \textbf{80.14} & \textbf{82.15} \\ [3pt] 
 & dt\_gbm & \unexpcol{-0.01} & 42.42 & \textbf{68.13} & 44.38 & \textbf{17.39} & \textbf{10.32} & 1.82 & \textbf{1.44} & \textbf{0.79} & \textbf{0.64} & 0.41 & 0.12 & \textbf{0.00} & \unexpcol{-0.02} & \textbf{0.34} \\  
 & dt\_rf & \textbf{0.00} & \textbf{52.54} & 57.10 & \textbf{44.61} & 16.63 & 6.19 & \textbf{2.19} & 0.96 & 0.51 & 0.00 & \textbf{0.48} & \textbf{0.33} & 0.00 & \textbf{0.00} & 0.10 \\  
\midrule
senseit\_aco & lpm\_gbm & 173.71 & 170.68 & 63.95 & \textbf{44.20} & 33.49 & 22.99 & 19.14 & 13.50 & 10.29 & 7.59 & 6.26 & 5.92 & \textbf{5.30} & 4.89 & 4.32 \\  
 & lpm\_rf & \textbf{177.67} & \textbf{181.26} & \textbf{79.86} & 42.86 & \textbf{37.60} & \textbf{28.80} & \textbf{23.75} & \textbf{19.06} & \textbf{13.91} & \textbf{10.74} & \textbf{8.48} & \textbf{6.09} & 5.20 & \textbf{5.32} & \textbf{4.62} \\ [3pt] 
 & dt\_gbm & 14.89 & 0.00 & \textbf{3.71} & 2.32 & \textbf{4.85} & 0.81 & \textbf{0.00} & - & - & - & - & - & - & - & - \\  
 & dt\_rf & \textbf{20.03} & \textbf{2.54} & 3.64 & \textbf{5.91} & 3.34 & \textbf{2.63} & 0.00 & 0.00 & - & - & - & - & - & - & - \\  
\midrule
senseit\_sei & lpm\_gbm & 160.59 & \textbf{65.27} & 23.44 & 10.48 & 6.76 & 4.86 & 4.82 & 4.46 & 4.79 & 4.12 & 4.54 & \textbf{5.17} & 3.91 & 4.21 & \textbf{4.46} \\  
 & lpm\_rf & \textbf{165.98} & 63.72 & \textbf{31.58} & \textbf{14.94} & \textbf{9.07} & \textbf{5.79} & \textbf{4.95} & \textbf{5.07} & \textbf{5.24} & \textbf{4.70} & \textbf{4.60} & 3.74 & \textbf{4.30} & \textbf{4.35} & 4.35 \\ [3pt] 
 & dt\_gbm & \textbf{2.66} & \textbf{1.01} & \textbf{3.49} & \textbf{2.29} & \textbf{0.95} & \textbf{1.30} & \textbf{1.37} & 0.00 & - & - & - & - & - & - & - \\  
 & dt\_rf & 2.33 & 0.00 & 3.36 & 1.65 & 0.87 & 0.00 & \unexpcol{-1.23} & - & - & - & - & - & - & - & - \\  
\midrule
covtype & lpm\_gbm & \textbf{36.87} & \textbf{49.24} & \textbf{12.78} & \textbf{11.21} & 7.84 & 7.15 & 7.15 & 8.07 & 7.70 & 8.25 & \textbf{10.94} & \textbf{8.35} & 4.37 & 8.77 & 5.84 \\  
 & lpm\_rf & 32.15 & 39.49 & 10.49 & 8.53 & \textbf{8.11} & \textbf{8.59} & \textbf{9.61} & \textbf{11.99} & \textbf{11.22} & \textbf{9.91} & 8.47 & 8.16 & \textbf{10.34} & \textbf{13.76} & \textbf{12.92} \\ [3pt] 
 & dt\_gbm & 342.27 & 92.85 & 43.23 & \textbf{20.04} & 8.14 & \textbf{8.05} & \textbf{5.67} & 3.26 & \textbf{4.92} & \textbf{3.52} & \textbf{2.72} & 0.00 & \textbf{0.00} & \textbf{0.00} & \textbf{1.74} \\  
 & dt\_rf & \textbf{354.45} & \textbf{98.94} & \textbf{50.87} & 14.10 & \textbf{9.46} & 7.38 & 4.76 & \textbf{4.20} & 0.94 & 1.81 & 2.30 & \textbf{0.71} & \unexpcol{-0.37} & 0.00 & 0.00 \\  
\midrule
connect-4 & lpm\_gbm & \textbf{37.62} & 11.66 & 12.01 & 6.84 & 5.68 & 6.82 & 4.58 & 2.10 & 3.82 & 3.21 & \textbf{3.02} & \textbf{3.64} & \textbf{2.32} & \textbf{2.97} & \textbf{3.40} \\  
 & lpm\_rf & 33.77 & \textbf{12.99} & \textbf{17.60} & \textbf{14.66} & \textbf{15.91} & \textbf{10.73} & \textbf{6.38} & \textbf{5.35} & \textbf{7.07} & \textbf{6.98} & 2.84 & 3.14 & 2.09 & 2.52 & 2.46 \\ [3pt] 
 & dt\_gbm & 89.33 & \textbf{29.23} & 20.20 & \textbf{12.10} & 9.73 & 9.88 & 7.82 & 7.43 & 0.57 & 4.61 & 1.08 & \textbf{3.35} & 2.23 & \textbf{1.15} & \textbf{1.55} \\  
 & dt\_rf & \textbf{113.71} & 21.91 & \textbf{20.52} & 11.23 & \textbf{16.86} & \textbf{10.96} & \textbf{10.64} & \textbf{9.11} & \textbf{6.51} & \textbf{5.88} & \textbf{6.76} & 2.16 & \textbf{2.97} & 0.61 & 0.00 \\  
\bottomrule [1.1pt]
\end{tabular}} \end{table*}

\begin{figure*}[ht]
\centerline{\includegraphics[width=1.7\columnwidth]{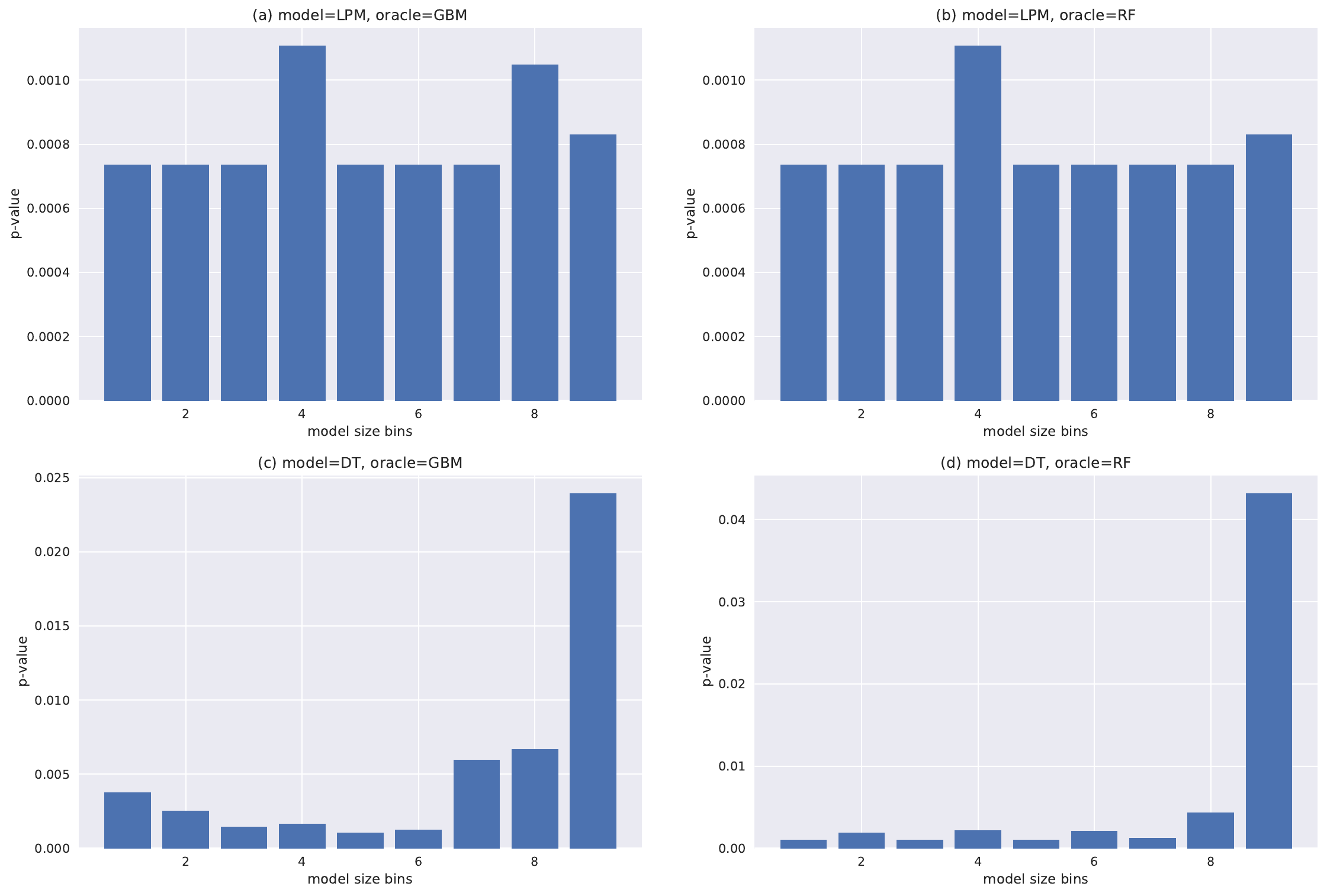}}
\caption{These plots show the \emph{p-values} for the Wilcoxon signed-rank test, with the null hypothesis $H_0$: using the oracle does \emph{not} produce better F1 test scores. The bin boundaries are selected using the \emph{Sturges} rule \protect\citep{sturges}.  Low \emph{p-values} favor our algorithm.}
\label{fig:stat_sig}
\end{figure*}

We also perform a \emph{Wilcoxon signed-rank test} \citep{wilcoxon} to measure  statistical significance. We use this test as it has been shown to be useful in comparing classifiers \citep{JMLR:v7:demsar06a,JMLR:v17:benavoli16a,japkowicz_shah_2011}. Results are shown in figure \ref{fig:stat_sig} for the following test setup:
\begin{enumerate}
    \item We divide the analysis by model size. This is because size strongly influences $\delta F1$ (as in Table \ref{tab:internal_improvements}).
    \item Normalized model sizes are used. Binning of model sizes is done using \emph{Sturges rule} \citep{sturges}.
    \item The \emph{one-sided} version of the \emph{paired} test is performed for each bin, where  pairs of scores $F1^{base}$ and $F1^*$ for a dataset, for models with sizes assigned to the bin, are compared. In cases were where multiple model sizes for a dataset fall within the same bin, $F1^{base}$ and $F1^*$ are first averaged and then compared. 
    \item The following hypotheses are tested:
    \begin{itemize}
        \item $\mathbf{H_0}$, null hypothesis: accuracies of models produced by our technique are not better.
        \item  $\mathbf{H_1}$, alternate hypothesis: accuracies of models trained using the oracle are better.
    \end{itemize}
    \emph{p-values} are shown for each bin. Small \emph{p-values} favor $\mathbf{H_1}$, i.e., our algorithm.
    \item Scores of $\delta F1=0$ are split equally between positive and negative ranks\footnote{The \texttt{zplit} option in \url{https://numpy.org/doc/stable/reference/generated/numpy.histogram_bin_edges.html} is used.}.
\end{enumerate}


\subsection{Multivariate Model Sizes}
\label{sec:vector_model_sizes_appendix}
As a continuation of the result shown in \S \ref{sec:vector_model_sizes}, we show results on the datasets (a) \emph{senseit-sei}  (b) \emph{higgs} (c) \emph{cod-rna} and (d) \emph{senseit-aco} here. We continue to observe pattern that as model sizes increase, in terms of both $max\_depth$ and $num\_trees$, improvements decrease.

\begin{figure}[h]
\centering
\includegraphics[width=0.95\columnwidth]{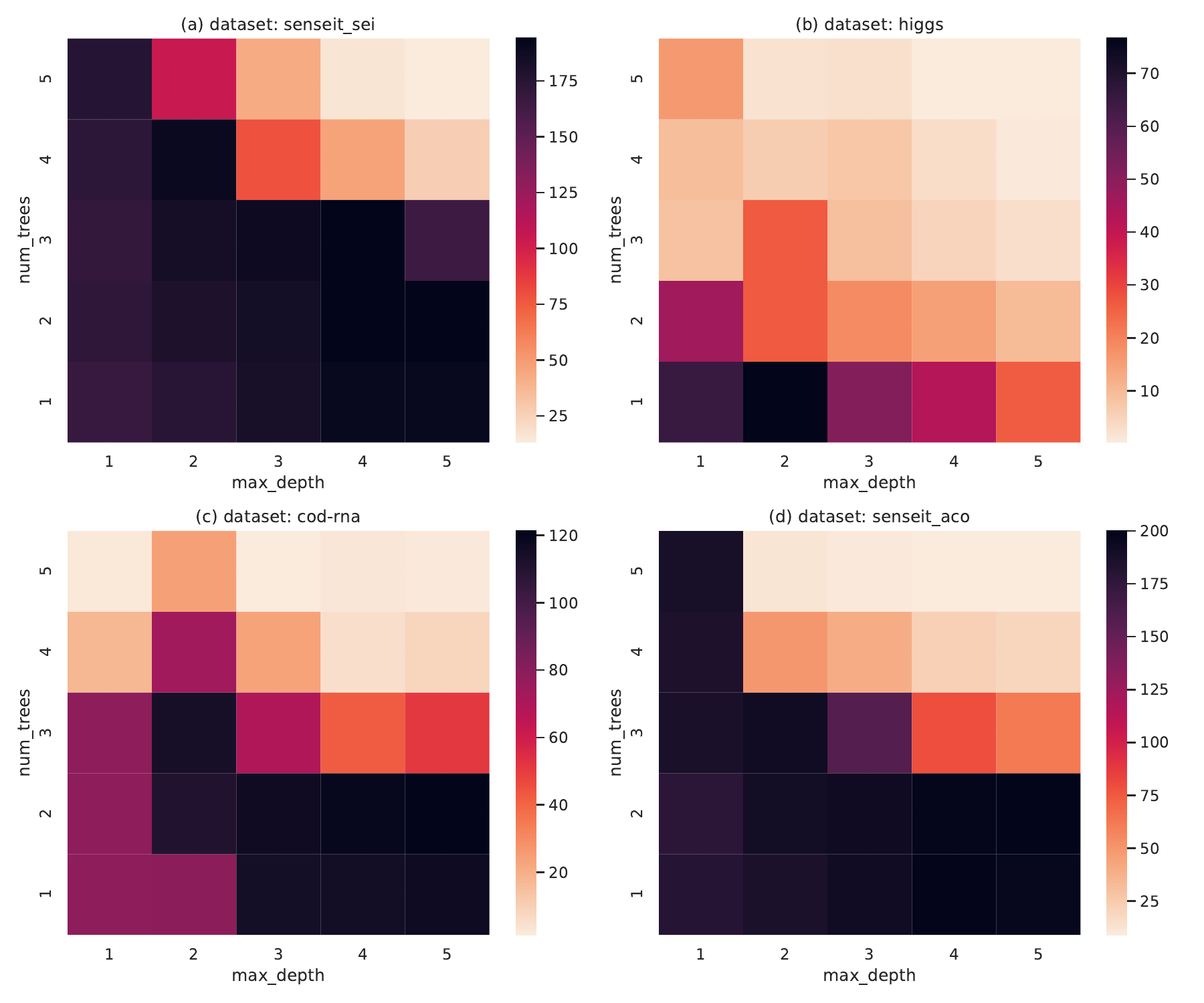}
\caption{Improvements in test $F1$-macro for multiple datasets for different sizes of $GBM$ models are shown. (a) Top-left: \emph{senseit-sei}  (b) Top-right: \emph{higgs} (c) Bottom-left:\emph{cod-rna} and (d) Bottom-right: \emph{senseit-aco}. Here, model size is the combination of \emph{max\_depth}  and \emph{number of trees} in the $GBM$ model. Greater improvements are seen at lower sizes.}
\label{fig:all_gbm_results}
\end{figure}

\subsection{Different Feature Spaces}
\label{sec:different_feature_space_appendix}
In our validation experiments in \S \ref{sec:val}, the feature vector representation was identical for the oracle and the interpretable model. This is also what Algorithm \ref{algo:main_with_data} implicitly assumes. Here, we consider the possibility of going a step further and using different feature vectors. If $f_{\mathcal{O}}$ and $f_{\mathcal{I}}$ are the feature vector creation functions for the oracle and the interpretable model respectively, and $x_i$ is a ``raw data'' instance, then:
\begin{enumerate}
    \item The oracle is trained on instances $f_{\mathcal{O}}(x_i)$, and provides uncertainties $u_{\mathcal{O}}(f_{\mathcal{O}}(x_i))$.
    \item The interpretable model is provided with data $f_{\mathcal{I}}(x_i)$, but the uncertainty scores available to it are $u_{\mathcal{O}}(f_{\mathcal{O}}(x_i))$. 
\end{enumerate}

The motivation for using different feature spaces is that the combination $(\mathcal{O}, f_{\mathcal{O}})$ may be known to work well together and/or a pre-trained oracle might be available only for this combination.

\begin{figure*}[h]
\centerline{\includegraphics[width=1.9\columnwidth]{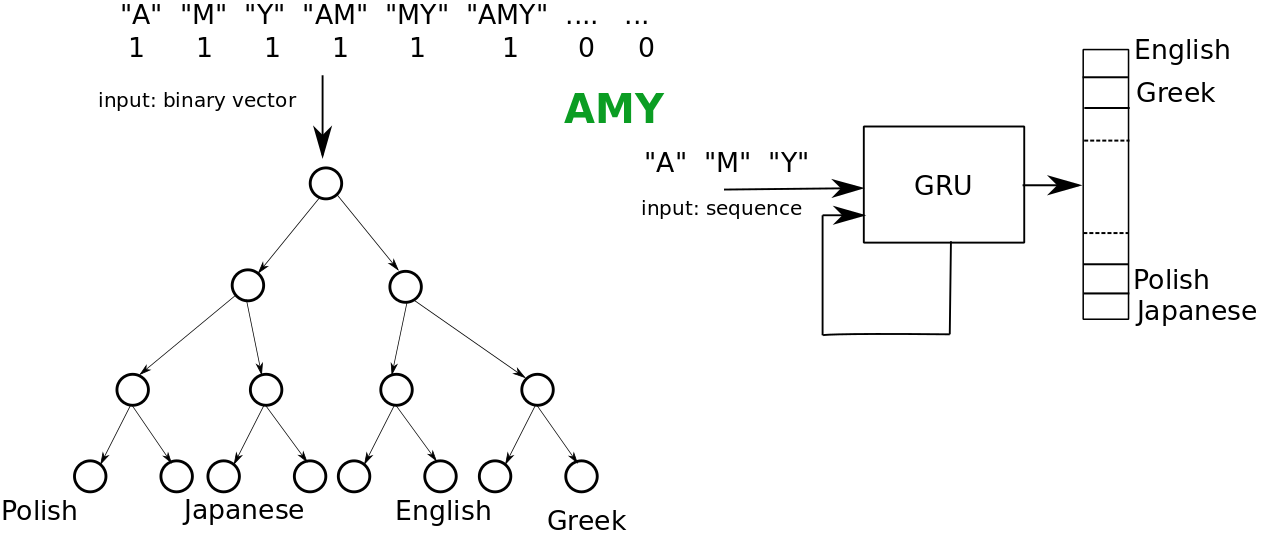}}
\caption[Different feature representations between the oracle and interpretable models.]{The feature representations for the oracle and the interpretable model may be different. Consider the name ``Amy'': the GRU is provided its letters, one at a time, in sequence, while the DT is given an n-gram representation of the name. }
\label{fig:rnn_dt_scehmatic}
\end{figure*}

We illustrate this application with the example of predicting nationalities from surnames of individuals. Our dataset \citep{NLP-Pytorch} contains examples from $18$ nationalities: \emph{Arabic, Chinese, Czech, Dutch, English, French, German, Greek, Irish, Italian, Japanese, Korean, Polish, Portuguese, Russian, Scottish, Spanish, Vietnamese}. The representations and models are as follows:
\begin{enumerate}
    \item The oracle model is a \emph{Gated Recurrent Unit (GRU)} \citep{cho-etal-2014-learning}, that is learned on the sequence of characters in a surname. The GRU is calibrated with \emph{temperature scaling} \citep{rnn_calibration}.
    \item The interpretable model is a DT, where the features are character n-grams, $n \in {1, 2, 3}$. The entire training set is initially scanned to construct an n-gram vocabulary, which is then used to create a sparse binary vector per surname - $1$s and $0$s indicating the presence and absence of an n-gram respectively.
\end{enumerate}
Figure \ref{fig:rnn_dt_scehmatic} shows a schematic of the setup.

\begin{figure}[h]
\centerline{\includegraphics[width=0.95\columnwidth]{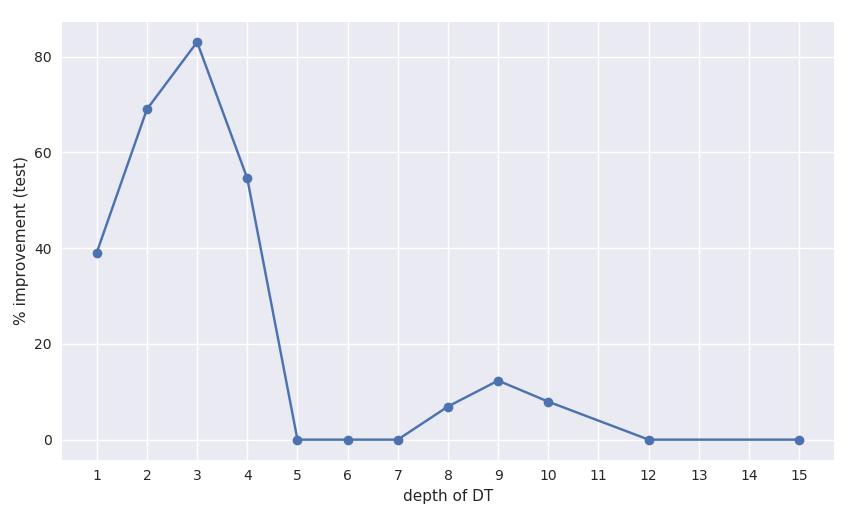}}
\caption{Improvements $\delta F1$ are shown for different depths of the DT. Reproduction of Figure \ref{fig:rnn_to_dt_improvements_main}, for convenience.}
\label{fig:rnn_to_dt_improvements}
\end{figure}

The n-gram representation leads to a vocabulary of $\sim 5000$ terms, that is reduced to $600$ terms based on a $\chi^2$-test in the interest of lower running time. DTs of different $depth \leq 15$ were trained. A budget of $T=3000$ iterations was used, and the relative improvement in the $F1$ macro score (as in Equation \ref{eqn:delta_f1}) is reported, averaged over three runs. Figure \ref{fig:rnn_to_dt_improvements} shows the results.

We see large improvements at small depths, that peak with $\delta F1=83.04\%$ at $depth=3$, and then again at slightly larger depths, which peak at $depth=9$ with $\delta F1=12.34\%$.

To obtain a qualitative idea of the changes in the DT using a oracle produces, we look at the prediction rules for \emph{Polish} surnames, when DT $depth=3$. For each rule, we also present examples of true and false positives.\\

\noindent \textbf{Baseline rules} - $precision=2.99\%, recall=85.71\%, F1=5.77\%$:
\begin{enumerate}[label=Rule \arabic*. , wide=0.5em,  leftmargin=*]
    \item  $k \wedge ski \wedge \lnot v $ 
    \begin{itemize}
        \item True Positives: \emph{jaskolski, rudawski}
        \item False Positives: \emph{skipper (English), babutski (Russian)}
    \end{itemize}
    \item   $k \wedge \lnot ski \wedge \lnot v$
    \begin{itemize}
        \item True Positives: \emph{wawrzaszek, koziol}
        \item False Positives: \emph{konda (Japanese), jagujinsky (Russian)}
    \end{itemize}
\end{enumerate}

\noindent \textbf{Oracle-based DT rules} - $precision=25.00\%, recall=21.43\%, F1=23.08\%$:
\begin{enumerate}[label=Rule \arabic*. , wide=0.5em,  leftmargin=*]
    \item  $ski \wedge \lnot (b \lor  kin)$ 
    \begin{itemize}
        \item True Positives: \emph{jaskolski, rudawski}
        \item False Positives: \emph{skipper (English), aivazovski (Russian)}
    \end{itemize}
\end{enumerate}

We note that the baseline rules are in conflict w.r.t. the literal ``ski'', and taken together, they simplify to $k \wedge \lnot v$. This makes them extremely permissive, especially \emph{Rule 2}, which requires the literal ``k'' while needing ``ski'' and ``v'' to be absent. Not surprisingly, these rules have high recall ($=85.71\%$) but poor precision ($=2.99\%$), leading to $F1=5.77\%$.

In the case of the oracle-based DT, now we have only one rule, that requires the atypical trigram ``ski''. This improves precision ($= 25\%$), trading off recall ($= 21.43\%$), for a significantly improved $F1=23.08\%$.

\begin{figure}[h]
\centerline{\includegraphics[width=0.95\columnwidth]{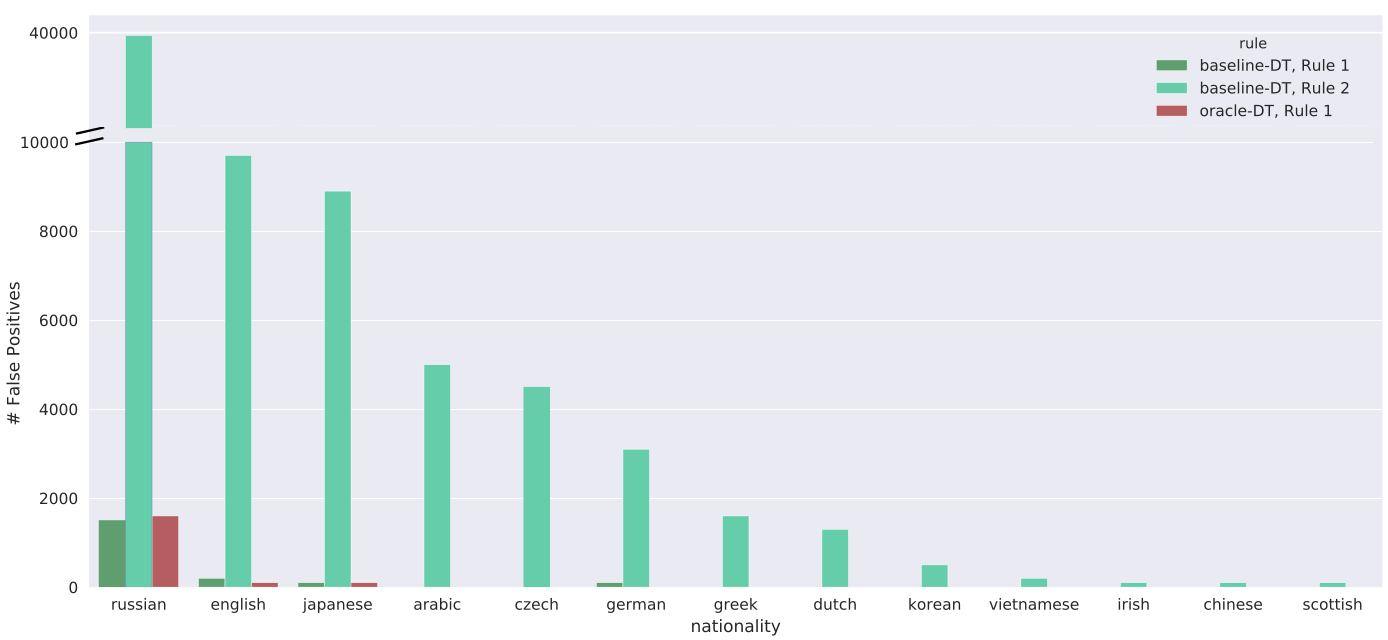}}
\caption{The distribution of nationalities in false positive predictions for the baseline and oracle based models, shown for predicting \emph{Polish} names. Only nationalities with non-zero counts are shown.
}
\label{fig:false_positives}
\end{figure}

The difference in rules may also be visualized by comparing the distribution of nationalities represented in their false positives, as in Figure \ref{fig:false_positives}. We see that the baseline DT rules, especially \emph{Rule 2}, predict many nationalities, but in the case of the DT learned using the oracle, the model confusion is concentrated around \emph{Russian} names, which is reasonable given the shared \emph{Slavic} origin of many \emph{Polish} and \emph{Russian} names.

We believe this is a particularly powerful and exciting application of our technique, and opens up a wide range of possibilities for translating information between models of varied capabilities.

\end{document}